%% file: neurips_2026.tex
\documentclass{article}

\usepackage[preprint]{neurips_2026}

\usepackage[utf8]{inputenc} % allow utf-8 input
\usepackage[T1]{fontenc}    % use 8-bit T1 fonts
\usepackage{hyperref}       % hyperlinks
\usepackage{url}            % simple URL typesetting
\usepackage{booktabs}       % professional-quality tables
\usepackage{amsmath}        % advanced math environments
\usepackage{amsfonts}       % blackboard math symbols
\usepackage{amssymb}        % additional math symbols
\usepackage{nicefrac}       % compact symbols for 1/2, etc.
\usepackage{microtype}      % microtypography
\usepackage{xcolor}         % colors
\usepackage{enumitem}       % customizable lists
\usepackage{multirow}       % multirow cells in tables
\usepackage{algorithm}      % algorithm environment
\usepackage{algpseudocode}  % algorithmic pseudocode
\usepackage{amsthm}
\newtheorem{assumption}{Assumption}
\newtheorem{remark}{Remark}
\usepackage{pifont}
\usepackage{graphicx} 
\usepackage[table]{xcolor} 
\definecolor{citecolor}{HTML}{0071bc}
\hypersetup{
    colorlinks=true,
    linkcolor=red,   
    citecolor=citecolor,  
    urlcolor=magenta!70!black  
}
\newcommand\blfootnote[1]{%
  \begingroup\renewcommand\thefootnote{}\footnote{#1}\addtocounter{footnote}{-1}\endgroup
}

\definecolor{uclablue}{RGB}{159, 195, 224}

\definecolor{uclagold}{RGB}{254,180,167}

\definecolor{grayred}{RGB}{232,237,205}

\title{Sketch Then Paint: \\Hierarchical Reinforcement Learning for Diffusion Multi-Modal Large Language Models}

\author{%
  \textbf{Siqi Luo}\textsuperscript{1,2,*},
  \textbf{Jianghan Shen}\textsuperscript{2,3,*},
  \textbf{Yi Xin}\textsuperscript{3,4},
  \textbf{Huayu Zheng}\textsuperscript{1},
  \textbf{Haoxing Chen}\textsuperscript{3},
  \textbf{Yan Tai}\textsuperscript{1},
  \textbf{Qi Qin}\textsuperscript{2},\\[0.2em]
  \textbf{Yue Li}\textsuperscript{5},
  \textbf{Junjun He}\textsuperscript{2,4},
  \textbf{Yihao Liu}\textsuperscript{2},
  \textbf{Guangtao Zhai}\textsuperscript{1},
  \textbf{Yuewen Cao}\textsuperscript{2},
  \textbf{Xiaohong Liu}\textsuperscript{1,4,\dag}\\[0.5em]
  \small
  \textsuperscript{1}Shanghai Jiao Tong University,~
  \textsuperscript{2}Shanghai Artificial Intelligence Laboratory,~
  \textsuperscript{3}Nanjing University,~\\
  \textsuperscript{4}Shanghai Innovation Institute,~
  \textsuperscript{5}Peking University%
  \\
  \raisebox{-0.3em}{\includegraphics[height=1.44em]{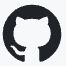}} \textbf{Code:}  \url{https://github.com/Alpha-VLLM/Lumina-DiMOO}
}
\begin{document}

\maketitle

\blfootnote{$^\ast$Equal contribution \quad $^\dagger$Corresponding author: \texttt{xiaohongliu@sjtu.edu.cn}}

\input{sec/0_abstract}

\input{sec/1_intro}
\input{sec/2_preliminary}
\input{sec/3_method}

\input{sec/4_experiments}
\input{sec/6_conclusion}

\clearpage
\bibliographystyle{unsrt}
\bibliography{references}

\clearpage
\appendix
\input{sec/A_supplementary}

% \clearpage
% \input{sec/X_checklist}

\end{document}

%% file: sec/0_abstract.tex
\vspace{-0.2in}
\begin{abstract}
    % Applying reinforcement learning (RL) to align Diffusion Multi-Modal Large Language Models (dMLLMs) for image generation is challenging because the same image can arise through multiple valid unmasking orderings, making importance ratios intractable. 
    Diffusion Multi-Modal Large Language Models (dMLLMs) are powerful for image generation, but optimizing them through reinforcement learning (RL) remains a major challenge.
    One primary difficulty is that a single image can be generated through many different unmasking sequences, which makes calculating importance ratios is often intractable.
    Additionally, existing methods tend to ignore the hierarchical generation process of dMLLMs, where early tokens define the global layout and later tokens focus on local details. 
    By assigning uniform rewards to all tokens, these current methods fail to reflect the actual contribution of each token to the final image.
    %
    % the unmasking process is inherently hierarchical, with early tokens determining global layout under high uncertainty and late tokens refining local details. 
    %
    % Consequently, random remasking mixes structural and refinement tokens and exposes invisible tokens, trajectory recording binds every update to a single ordering, and both uniformly broadcast image-level rewards ignoring their fundamentally different contributions.
    %
    To address these issues, we propose \underline{\textbf{H}}ierarchical
    \underline{\textbf{T}}oken \underline{\textbf{GRPO}} (\textbf{HT-GRPO}), which integrates this hierarchy directly into the policy optimization process.
    Our approach features a Sketch-Then-Paint training scheme that organizes updates into three distinct stages: global, structure, and refinement.
    %
    % We also employ a prompt-conditioned estimator to compute importance ratios from a fully masked state.
    We also use a prompt-conditioned estimator to calculate importance ratios starting from a fully masked state.
    %
    % Furthermore, our Hierarchical Credit Assignment mechanism prioritizes critical structural tokens to ensure accurate reward propagation.
    Furthermore, we introduce a Hierarchical Credit Assignment mechanism that prioritizes key structural tokens to ensure accurate reward propagation. 
    Experiments using two popular dMLLM backbones, MMaDA and Lumina-DiMOO, demonstrate that HT-GRPO achieves substantial gains on the GenEval and DPG benchmarks.
    Evaluations across six additional metrics confirm significant improvements in image quality, aesthetics, and human preference.
    
    % \textbf{Sketch-Then-Paint} staged training organizes updates into \textbf{Global}, \textbf{Structure}, and \textbf{Refinement} stages by decoding order, covering multiple generation paths via random subset sampling; a prompt-conditioned estimator computes importance ratios under a fully masked state, eliminating contamination without caching intermediate states. Hierarchical Credit Assignment assigns higher credit to structural tokens and lower credit to refinement tokens, aligning reward propagation with actual contributions.
    % Experiments show HT-GRPO consistently improves two dMLLM backbones: MMaDA improves from 0.56 to 0.83 on GenEval and from 70.51 to 81.09 on DPG-Bench; Lumina-DiMOO improves from 0.83 to 0.92 on GenEval and from 81.86 to 84.47 on DPG-Bench.
    \end{abstract}

%% file: sec/1_intro.tex
\section{Introduction}
\label{sec:introduction}
Recent advancements in text-to-image (T2I) generation, driven by powerful diffusion models~\cite{cai2025z,qin2025lumina,wu2025qwen,team2025longcat} and autoregressive (AR) models~\cite{wang2024emu3,cui2025emu35nativemultimodalmodels,xin2025lumina,liu2026lumina}, have achieved significant breakthroughs. 
This success is largely built upon the scaling of computational resources, model parameters, and training data. 
Alongside these efforts, the integration of reinforcement learning (RL)~\cite{liu2025flow,xue2025dancegrpo,geng2025x,yuan2025argrpo} has also become an essential component. 
By effectively aligning generation outputs with human preferences (e.g., prompt adherence, aesthetic), RL significantly improves the performance of T2I models. 
More recently, Diffusion Multi-Modal Large Language Models (dMLLMs)~\cite{mmada,luminadimoo,muddit,LLaDA2Uni,li2025lavida,you2026lladao} have emerged as a significant architectural evolution. 
By performing parallel iterative denoising over discrete tokens, dMLLMs offer a novel approach to visual generation. 
This architectural shift naturally requires researchers to design RL strategies specifically tailored for dMLLMs.
The generation process of dMLLMs is inherently hierarchical: tokens decoded
early shape global layout under high uncertainty, while tokens decoded late
refine local details under low uncertainty.

\begin{figure}[t]
    \centering
    \vspace{-0.1in}
    \includegraphics[width=1\linewidth]{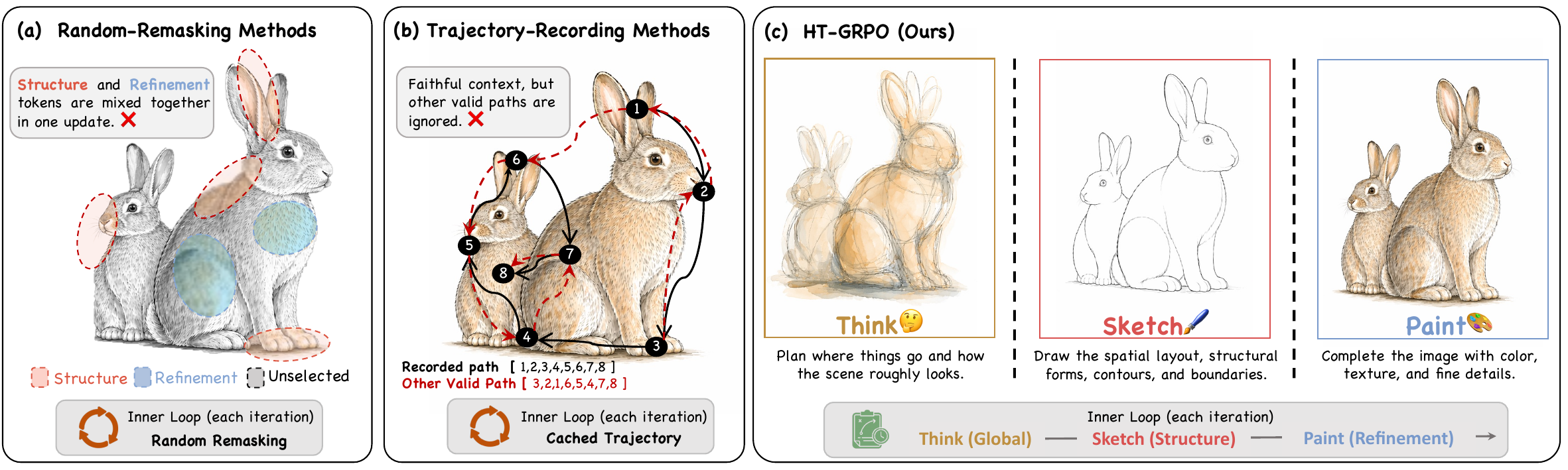}
    \vspace{-0.2in}
      \caption{\textbf{Comparison of RL inner-loop strategies for dMLLMs.}
          % (a) Random remasking methods mix structural and refinement tokens
          % in the same gradient step.
          % (b) Trajectory recording methods bind all updates to a single
          % generation ordering.
          % (c) HT-GRPO organizes updates into three hierarchical stages (Global$\to$Structure$\to$Refinement) with random subset sampling.
          }               
    \label{fig:insight}
    \vspace{-0.2in}
\end{figure}

Existing RL approaches generally follow two routes, but both fail to account
for this structure: 
\textbf{1) Random remasking methods} (shown in Figure~\ref{fig:insight}\textcolor{red}{(a)}) ~\cite{maskgrpo,d1,unigrpo} approximate intermediate states by randomly remasking tokens in generated images and evaluating selected tokens under the remaining visible context (\underline{\textit{Limitation: future-token contamination}}).
However, this earliest naive paradigm leaks future information into the selected step.
It also arbitrarily mixes structural tokens responsible for global layout with refinement tokens responsible for local details into the same gradient update, preventing the model from optimizing each group at its appropriate stage (\underline{\textit{Limitation: stage conflation}}).
%
% It also disrupts the naturally staged generation process, incorrectly mixing the high-entropy gradients used for overall composition with the low-entropy gradients used for fine textures.
%
% However, these methods have two main drawbacks.
% %
% First, they often retain tokens that should be invisible.
% %
% This leaks future information into the current step and leads to future-token contamination. 
% %
% Second, they mix tokens from different generation stages within the same optimization process.
% %
% \textcolor{red}{This disrupts the naturally staged generation process, incorrectly mixing the high-entropy gradients used for overall composition with the low-entropy gradients used for fine textures.}
%
\textbf{2) Trajectory recording methods} (shown in Figure~\ref{fig:insight}\textcolor{red}{(b)}) ~\cite{tracerl,cjgrpo,dtreerpro,agrpo} build a more explicit decision process by storing actual denoising trajectory.
While this prevents information leakage, the inner-loop updates for each generated image are bound to a single recorded ordering, ignoring the many other valid unmasking orderings (\underline{\textit{Limitation: narrow path coverage}}).    
Beyond these specific issues, both routes assign uniform image-level rewards to all tokens, ignoring the fact that different tokens have entirely different impacts on the final image quality (\underline{\textit{Shared Limitation: uniform token reward}}). 
% %
% In fact, the generation process of dMLLMs is inherently hierarchical, as shown in Figure~\ref{fig:insight}.
% %
% % Early tokens determine structural with high uncertainty like layout, while later tokens focus on refining details like color. 
% %
% Treating all tokens equally fails to reflect this underlying mechanism.

To overcome these limitations, we propose Hierarchical
Token GRPO (HT-GRPO), a novel
RL method that explicitly encodes the hierarchical structure of dMLLM generation
into inner-loop policy optimization.
HT-GRPO consists of two core components.
\textbf{First, we propose a Sketch-Then-Paint staged training framework.}
This organizes the inner-loop updates into three consecutive stages (shown in Figure~\ref{fig:insight}\textcolor{red}{(c)}): a Global stage that updates all tokens jointly to establish a stable coarse foundation, a Structure stage that focuses exclusively on structural tokens to sharpen global composition, and a Refinement stage that polishes local details only after the structure is settled, preventing structural and refinement tokens from being mixed in the same gradient update.
Within each stage, random subset sampling covers multiple valid generation paths at the same semantic level. This enables the model to explore and optimize over a rich set of semantically consistent generation paths.
To support this staged optimization, we further introduce a \textbf{prompt-conditioned estimator} that computes token-level importance ratios under a unified fully masked state using only the text prompt as condition.
This simultaneously eliminates future-token contamination, prevents low-entropy degradation when evaluating refinement tokens, and removes the need to cache any intermediate denoising states.
\textbf{Second, we introduce Hierarchical Credit Assignment.}
This assigns higher credit weights to early structural tokens and lower weights
to later refinement tokens, ensuring that image-level rewards are distributed in proportion to each token's actual contribution.

% to the final image quality.
Extensive experiments confirm that HT-GRPO delivers superior text-image alignment on multiple benchmarks.
On the MMaDA model, it raises GenEval from 0.56 to 0.83 and DPG from 70.51 to 81.09.
When applied to Lumina-DiMOO, these scores further improve to 0.92 and 84.47 respectively.
Additionally, HT-GRPO significantly boosts performance across six metrics covering image quality, aesthetics, and human preference.
These results confirm that hierarchical token optimization is a versatile and effective approach. 
Overall, our contributions can be summarized as follows:

% , and additional 6 metrics for human preference and visual quality.
% Extensive experiments on MMaDA~\cite{mmada} and Lumina-DiMOO~\cite{luminadimoo} show that HT-GRPO consistently improves text-image alignment across GenEval~\cite{geneval}, DPG-Bench~\cite{dpgbench}, and 6 metrics for human preference, visual quality and aesthetics. 
% %
% HT-GRPO improves GenEval from 0.56 to 0.83 and DPG from 70.51 to 81.09 on MMaDA and GenEval from 0.83 to 0.92 and DPG from 81.86 to 84.47 on Lumina-DiMOO, respectively. 
% These results demonstrate the generality of hierarchical token optimization.
% Our main contributions are as follows:
\vspace{-0.1in}
\begin{itemize}[leftmargin=*]
\item We reveal that the intrinsic hierarchical structure of dMLLM generation is essential. Existing RL methods ignore it, leading to stage conflation and inaccurate reward broadcasts. Instead, we explicitly leverage this structure to achieve principled hierarchical policy optimization.

% We identify the intrinsic hierarchical structure of dMLLM generation as a double-edged sword: ignored by existing methods, it causes stage conflation, future-token contamination, and a credit assignment problem from uniform reward broadcast; explicitly exploited, it enables principled hierarchical policy optimization.

\item We propose \textbf{Sketch-Then-Paint} staged training, which divides inner-loop updates into Global, Structure, and Refinement stages by generation order, and uses a prompt-conditioned estimator to prevent future-token contamination without caching trajectories.

% organizes inner-loop updates into Global, Structure, and Refinement stages by generation order, with a prompt-conditioned estimator that eliminates future-token contamination and low-entropy degradation without caching trajectories.

\item We design \textbf{Hierarchical Credit Assignment} to allocate generation-order-aware weights to different tokens. This effectively aligns the reward propagation with the actual contribution of each token.

% , assigning generation-order-aware weights to structural and refinement tokens to align reward propagation with each token's actual contribution.

\item We show that HT-GRPO consistently enhances MMaDA and Lumina-DiMOO across various benchmarks. This confirms the strong generalization of hierarchical token optimization across dMLLM architectures, outperforming leading T2I models.

% HT-GRPO consistently improves both MMaDA and Lumina-DiMOO across compositional, aesthetic, and human-preference benchmarks, demonstrating that hierarchical token optimization generalizes across dMLLM architectures
% and surpasses leading T2I models.
\end{itemize}

%% file: sec/2_preliminary.tex
\begin{figure}[t]
    \centering
    \vspace{-0.05in}
    \includegraphics[width=1\linewidth]{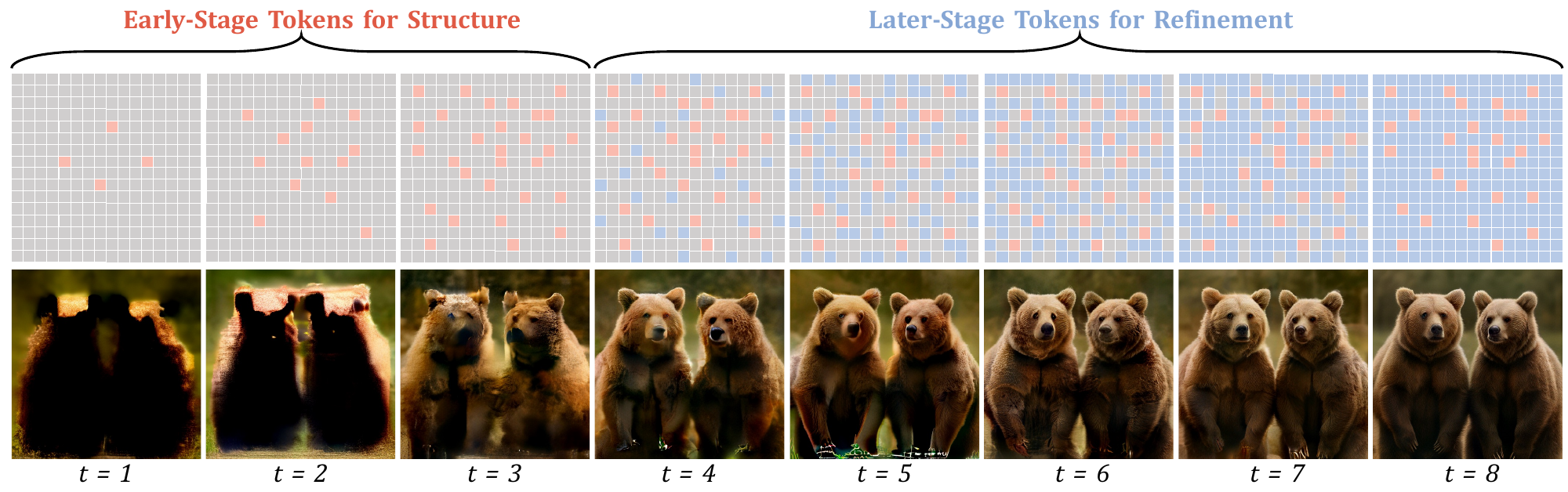}
    \vspace{-0.2in}
      \caption{\textbf{Visualization of the image generation process in dMLLMs.} 
      Top: early (red, structure) tokens set layout, later (blue, refinement) tokens add detail.
      Bottom: depict the corresponding outputs.}              
    \label{fig:insight_preliminary}
    \vspace{-0.1in}
\end{figure}

\vspace{-0.1in}
\section{Preliminaries}
\label{sec:preliminaries}

\subsection{Image Generation Process of dMLLMs}
\label{subsec:dmlm}

dMLLMs generate images by iteratively denoising masked visual tokens.
An image is represented as a token sequence of length $N$ with text condition $c$.
Starting from a fully masked state $\mathbf{x}^{(T)}$, the model repeatedly unmasks a subset of positions until the final image $\mathbf{x}^{(0)}=(v_1,\ldots,v_N)$ is produced:
\begin{equation}
\label{eq:dmlm_denoise}
\mathbf{x}^{(t-1)} \sim \pi_\theta\!\left(\cdot \mid \mathbf{x}^{(t)}, c\right),
\qquad t=T,T-1,\ldots,1.
\end{equation}
Unlike autoregressive models, which generate tokens in a fixed left-to-right order, dMLLMs can produce the same image through many different unmasking orderings.
We track this ordering with the generation-order rank $\rho_{g,i} \in \{1,\ldots,N\}$: rank 1 is the first position unmasked and rank $N$ the last.
Tokens unmasked in the same step receive consecutive ranks under an arbitrary fixed order within that step.

The rank $\rho_{g,i}$ reflects a fundamental asymmetry in what the model knows at prediction time.
A token with small $\rho_{g,i}$ is unmasked early, when most of the image is still masked.
It faces high uncertainty, and its choice largely determines where objects appear and how the scene is organized.
A token with large $\rho_{g,i}$ is unmasked late, after most of the image has been revealed.
It faces much lower uncertainty and mainly refines color, texture, and local detail.
We call tokens with small $\rho_{g,i}$ \emph{structural tokens} and tokens with large $\rho_{g,i}$ \emph{refinement tokens} (Figure~\ref{fig:insight_preliminary}).
\hyperref[prop:entropy-monotonicity]{Proposition~C.1} 
% in Appendix~\ref{app:theory} 
provides a formal characterization of this entropy gap.
This rank-induced asymmetry becomes a key challenge when applying RL to dMLLMs, as formalized in Section~\ref{subsec:grpo}.

\subsection{Group Relative Policy Optimization}
\label{subsec:grpo}

GRPO~\cite{deepseekmath} maximizes a clipped surrogate objective over $G$ rollouts sampled from a behavior policy $\pi_{\theta_{\mathrm{old}}}$, using per-token importance ratios and group-relative advantages; the standard formulation is recalled in Appendix~\ref{app:grpo_standard}.

\noindent\textbf{Applying GRPO to dMLLMs.}
Extending the token-level GRPO objective to dMLLMs rests on two properties of Eq.~\eqref{eq:dmlm_denoise}.
First, \emph{Markov chain decomposition}: along a fixed unmasking trajectory, the joint probability factorizes step-wise via the chain rule.
Second, \emph{within-step conditional independence}: conditioned on the current state $\mathbf{x}^{(t)}$ and text prompt $c$, the tokens unmasked at step $t$ are sampled independently across positions.
Together, these two properties enable a per-token formulation of the importance ratio in dMLLMs, analogous to the autoregressive case (Appendix~\ref{app:grpo_standard}).

\begin{assumption}[Stage-varying optimization]
\label{asmp:stage_order}
Because dMLLMs decode tokens across $T$ stages under progressively changing visual contexts, the per-step optimization objective is inherently non-uniform.
We therefore introduce an \emph{optimization support set} $\mathcal{M}_g^{(k)}$ and a \emph{conditioning context} $\mathbf{C}_{g,i}^{(k)}$ to characterize each gradient step $k$.
All methods discussed in this paper are special instantiations of this paradigm. A unified comparison is given in Appendix~\ref{app:unified_framework}.
\end{assumption}
Under Assumption~\ref{asmp:stage_order}, the objective at gradient step $k\in\{1,\ldots,K\}$ is:
\begin{align}
\label{eq:grpo_dmlm_k}
\mathcal{J}^{(k)}(\theta)
= \mathbb{E}_{c,\,\{\mathbf{x}^{(0)}_g\}\sim\pi_{\theta_{\mathrm{old}}}}\!\Bigg[
  &\frac{1}{G}\sum_{g=1}^{G}
  \frac{1}{|\mathcal{M}_g^{(k)}|}\!\sum_{i\in\mathcal{M}_g^{(k)}}
  \min\!\Big(r_{g,i}^{(k)}(\theta)\,A_g,\;
  \mathrm{clip}\!\left(r_{g,i}^{(k)}(\theta),1{-}\epsilon,1{+}\epsilon\right)A_g\Big) \notag\\
  &-\;\beta\,\mathbb{D}_{\mathrm{KL}}\!\left(\pi_\theta\,\|\,\pi_{\mathrm{ref}}\right)
\Bigg],
\end{align}
where the per-token importance ratio
\vspace{-0.05in}
\begin{equation}
\label{eq:per_token_ratio}
r_{g,i}^{(k)}(\theta)
= \frac{\pi_\theta\!\left(v_{g,i}\mid\mathbf{C}_{g,i}^{(k)},\,c\right)}
       {\pi_{\theta_{\mathrm{old}}}\!\left(v_{g,i}\mid\mathbf{C}_{g,i}^{(k)},\,c\right)}
\end{equation}
measures the relative likelihood of visual token $v_{g,i}$ under the current versus behavior policy, conditioned on the visual context $\mathbf{C}_{g,i}^{(k)}$ and text context $c$.
The group-relative advantage $A_g = (R_g - \mathrm{mean}(\{R_j\})) / (\mathrm{std}(\{R_j\}) + \delta)$ is a scalar broadcast uniformly to every token in rollout $g$.

\begin{remark}
\label{rem:intractable}
Each rollout produces one image $\mathbf{x}^{(0)}_g$ via a single unmasking trajectory.
The same image could have been generated through exponentially many alternative orderings, each inducing a different conditioning context $\mathbf{C}_{g,i}^{(k)}$ for every token.
The ideal objective averages Eq.~\eqref{eq:grpo_dmlm_k} over all valid trajectories:
\begin{equation}
\label{eq:ideal}
\mathcal{J}^{*}(\theta)
= \mathbb{E}_{\tau\sim\pi_{\theta_{\mathrm{old}}}}\!\left[\mathcal{J}_{\tau}(\theta)\right],
\end{equation}
where $\mathcal{J}_{\tau}$ denotes Eq.~\eqref{eq:grpo_dmlm_k} instantiated with $\mathcal{M}_g^{(k)}=\{1,\ldots,N\}$ and $\mathbf{C}_{g,i}^{(k)}$ set to the tokens unmasked before position $i$ under trajectory $\tau$.
This expectation is intractable in practice.
\end{remark}

\subsection{Limitations of Existing Methods}
\label{subsec:limitations}
As noted in Remark~\ref{rem:intractable}, $\mathcal{J}^{*}$ is intractable in practice. Existing methods therefore approximate it by making different choices of $\mathcal{M}_g^{(k)}$ and $\mathbf{C}_{g,i}^{(k)}$.
Random remasking methods construct diverse pseudo-trajectories through random masking, while trajectory recording methods condition on the single observed trajectory.
Each approximation carries its own limitations, analyzed below.
A unified formal comparison is provided in Appendix~\ref{app:unified_framework}.

\noindent\textbf{Random Remasking Methods.} MaskGRPO~\cite{maskgrpo}, D1~\cite{d1}, and UniGRPO~\cite{unigrpo} schedule $\mathcal{M}_g^{(k)}$ by progressively increasing the mask ratio across the $K$ gradient steps.
At each step $k$, the conditioning context $\mathbf{C}_{g,i}^{(k)}$ consists of the randomly retained visible tokens together with the text prompt $c$.
This construction introduces two limitations.
First, \emph{future-token contamination}: the randomly retained tokens are not ordered by generation rank. The context $\mathbf{C}_{g,i}^{(k)}$ may therefore include tokens not yet unmasked when position $i$ was originally generated, making it causally inconsistent, as shown in \hyperref[prop:contamination]{Proposition~C.2}.
Second, \emph{stage conflation}: the progressive mask schedule assigns tokens to $\mathcal{M}_g^{(k)}$ without regard to generation-order rank $\rho_{g,i}$. Structural tokens and refinement tokens are thus mixed in the same gradient update despite their fundamentally different entropy levels, as established in Section~\ref{subsec:dmlm} and formalized in \hyperref[prop:entropy-monotonicity]{Proposition~C.1}.
On the other hand, the random remasking strategy provides multi-path coverage.
By randomly varying the masked tokens across gradient steps, these methods expose the model to multiple partial views of the same generated image, providing broader context diversity than a single fixed trajectory.

\noindent\textbf{Trajectory Recording Methods.} TraceRL~\cite{tracerl}, CJ-GRPO~\cite{cjgrpo}, d-TreeRPO~\cite{dtreerpro}, and AGRPO~\cite{agrpo} set $\mathcal{M}_g^{(k)}$ to all $N$ token positions at every gradient step.
The conditioning context $\mathbf{C}_{g,i}^{(k)}$ is the set of tokens already unmasked just before position $i$ was revealed during the rollout trajectory.
By grounding the context in the actual generation order, these methods avoid both future-token contamination and stage conflation.
They introduce two limitations of their own, however.
First, \emph{limited path coverage}: as discussed in Remark~\ref{rem:intractable}, the ideal $\mathcal{J}^*$ requires averaging over all valid trajectories. 
Each rollout records only one ordering, so all alternatives remain unseen and the expectation is approximated by a single sample.
Second, computational overhead: recording the full denoising trajectory requires storing all intermediate states and performing $O(T)$ forward passes per rollout. This substantially increases both memory usage and compute cost per training cycle.

\noindent\textbf{Shared Limitation.}
Neither family accounts for the difference between structural and refinement tokens in the advantage assignment.
The scalar $A_g$ is broadcast uniformly to every token regardless of denoising stage, assigning equal credit to structural and refinement tokens despite their vastly different entropy levels.

%% file: sec/3_method.tex
\vspace{-0.05in}
\section{Method}
\label{sec:method}
The limitations in Section~\ref{subsec:limitations} fall into two categories.
The first concerns the approximation of $\mathcal{J}^*$: existing choices of $\mathcal{M}_g^{(k)}$ and $\mathbf{C}_{g,i}^{(k)}$ each introduce stage conflation, limited path coverage, or future-token contamination.
The second concerns reward attribution: the image-level advantage $A_g$ is broadcast uniformly, ignoring the entropy gap between structural and refinement tokens.

HT-GRPO addresses both categories through two components.
Section~\ref{subsec:ht-grpo} presents Sketch-Then-Paint staged training with a prompt-conditioned estimator.
Inter-stage partitioning of $\mathcal{M}_g^{(k)}$ resolves stage conflation, intra-stage random subset sampling alleviates limited path coverage, and fixing $\mathbf{C}_{g,i}^{(k)} = \mathbf{C}_\emptyset$ eliminates future-token contamination.
Section~\ref{subsec:credit} replaces the uniform $A_g$ with Hierarchical Credit Assignment.

\begin{figure}
    \centering
    \vspace{-0.1in}
    \includegraphics[width=1\linewidth]{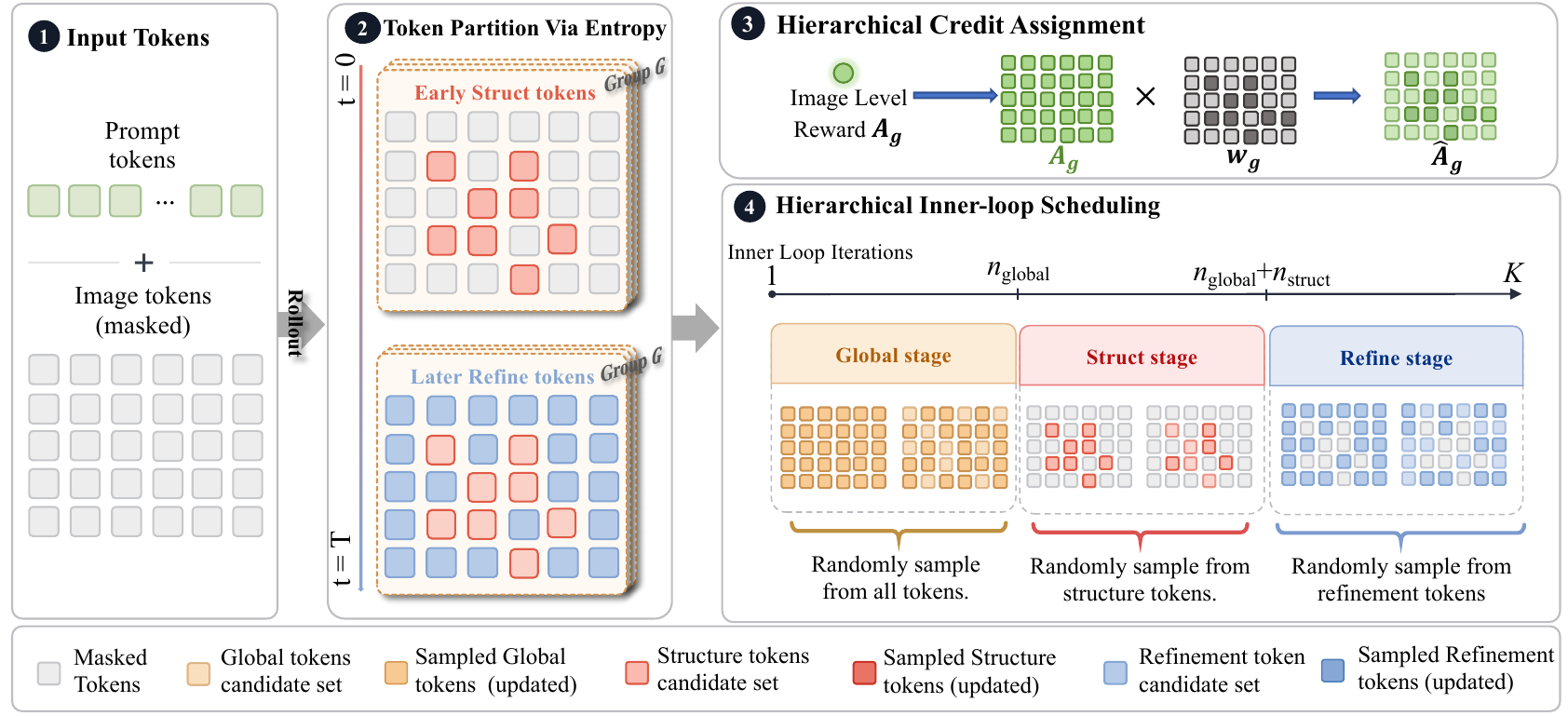}
    \vspace{-0.22in}
    \caption{\textbf{HT-GRPO framework.}
    (1) Prompt and masked image tokens are rolled out to produce $G$ sample groups.
    (2) Tokens are partitioned into structural (early, high-entropy) and
    refinement (later, low-entropy) sets via generation-order rank.
    (3) The image-level advantage $A_g$ is reweighted by per-token credits $w_{g,i}$
    to form the hierarchical advantage  $\tilde{A}_{g,i}$.
    (4) Inner-loop updates are scheduled into three stages, each randomly sampling from its corresponding token set.}
    \label{fig:method}
    \vspace{-0.06in}
\end{figure}

\vspace{-0.05in}
\subsection{Hierarchical Token GRPO}
\label{subsec:ht-grpo}

Visual generation is a progressive process from global structure to local detail, encoded in the distribution of $\rho_{g,i}$.
As established in Section~\ref{subsec:dmlm}, tokens with small $\rho_{g,i}$ are structural tokens facing high uncertainty, while those with large $\rho_{g,i}$ are refinement tokens operating under rich context.
HT-GRPO encodes this ordering into both the optimization schedule and the importance ratio estimator.

\subsubsection{Sketch-Then-Paint Staged Training}

Under standard GRPO, one policy optimization cycle is:
\begin{equation}
\pi_{\theta_{\mathrm{old}}}
\xrightarrow{\text{rollout}}
\bigl\{\mathbf{x}^{(0)}_g\bigr\}_{g=1}^{G}
\xrightarrow{\text{optimize}}
\theta
\xrightarrow{\text{synchronize}}
\theta_{\mathrm{old}}\leftarrow\theta.
\end{equation}
The policy performs $K$ inner-loop gradient updates on the same rollout batch.
Although some methods vary which tokens enter each update, none account for the entropy-level distinction between structural and refinement tokens.

We organize the $K$ inner-loop updates into three consecutive stages:
\begin{equation}
\text{Global} \rightarrow \text{Structure} \rightarrow \text{Refinement}.
\end{equation}
This creates a coarse-to-fine curriculum from high entropy to low entropy.
The Global stage first optimizes all tokens jointly to provide a stable starting point.
The Structure stage then focuses on high-entropy tokens, and the Refinement stage finally optimizes low-entropy detail after the skeleton has been established.
This progression reduces gradient interference between token groups with fundamentally different entropy regimes.

Let $N_s = \lfloor \alpha N \rfloor$ with $\alpha \in (0,1)$ denoting the structure fraction.
We define the stage-specific token sets as
\begin{equation}
\mathcal{S}_{g,\mathrm{global}} = \{1,\ldots,N\},
\mathcal{S}_{g,\mathrm{structure}} = \{i \mid \rho_{g,i} \le N_s\}, 
\mathcal{S}_{g,\mathrm{refinement}} = \{i \mid \rho_{g,i} > N_s\}.
\end{equation}
Here $N_s$ is a token-count cutoff rather than a denoising-step cutoff.
$\mathcal{S}_{g,\mathrm{structure}}$ collects the $N_s$ structural tokens with the smallest $\rho_{g,i}$ values, and $\mathcal{S}_{g,\mathrm{refinement}}$ the remaining $N - N_s$ refinement tokens, following the taxonomy in Section~\ref{subsec:dmlm}.
This partition depends only on the rollout's own unmasking order and requires no external annotation.
Using unmasking rank as an entropy proxy is justified by \hyperref[prop:entropy-monotonicity]{Proposition~C.1}.

We split the total inner-loop budget as $K = n_{\mathrm{global}} + n_{\mathrm{structure}} + n_{\mathrm{refinement}}$.
For the $k_s$-th update inside stage $s$, where $k_s = 0,\ldots,n_s-1$, we independently sample a random subset $\mathcal{M}_g^{(k_s)} \subseteq \mathcal{S}_{g,s}$ according to an annealed sampling rate $\gamma_{k_s}^{(s)}$.
Only tokens in $\mathcal{M}_g^{(k_s)}$ contribute gradients in that update:
\begin{equation}
\label{eq:schedule}
\gamma_{k_s}^{(s)} =
\gamma_{\min}^{(s)}
+
\bigl(\gamma_{\max}^{(s)} - \gamma_{\min}^{(s)}\bigr)
\frac{\max(1, n_s - 1) - k_s}{\max(1, n_s - 1)}.
\end{equation}
The sampling rate decays linearly from $\gamma_{\max}^{(s)}$ to $\gamma_{\min}^{(s)}$.
Early updates cover more tokens and provide stable gradient directions, while later updates reduce computation and sharpen focus.
Sampling different subsets $\mathcal{M}_g^{(k_s)}$ across inner-loop updates ensures that distinct portions of the token space receive gradient signal, rather than committing to a fixed subset throughout optimization.
This intra-stage sampling acts as a Monte Carlo estimator over the token space within each stage, partially alleviating the limited path coverage identified in Section~\ref{subsec:limitations}.

\subsubsection{Prompt-Conditioned Estimator}
\label{subsec:prompt-estimator}

Random remasking methods set $\mathbf{C}_{g,i}^{(k)}$ to a randomly retained subset of the generated image, which may expose tokens not yet visible at generation time and causes future-token contamination, as shown in \hyperref[prop:contamination]{Proposition~C.2}.
Trajectory recording methods avoid contamination by using the actual trajectory state, but require storing all intermediate denoising states and performing $O(T)$ forward passes per rollout cycle.

We instead use the fully masked initial state $\mathbf{x}^{(T)}$ as a unified conditioning context, denoted $\mathbf{C}_\emptyset$ to emphasize that all $N$ positions are masked, and define the prompt-conditioned importance ratio as
\begin{equation}
\label{eq:prompt_ratio}
r_{g,i}(\theta) =
\frac{
\pi_\theta\!\left(v_{g,i} \mid \mathbf{C}_\emptyset, c\right)
}{
\pi_{\theta_{\mathrm{old}}}\!\left(v_{g,i} \mid \mathbf{C}_\emptyset, c\right)
}.
\end{equation}
Under a fully masked input, dMLLMs predict all $N$ token distributions in a single forward pass, so all probabilities are obtained without caching any intermediate states.
Using $\mathbf{C}_\emptyset$ eliminates future-token contamination by construction, preserves predictive entropy for refinement tokens as shown in \hyperref[prop:entropy-lower-bound]{Proposition~C.3}, and avoids the memory cost of trajectory recording.

\subsection{Hierarchical Credit Assignment}
\label{subsec:credit}

Both families of existing methods uniformly broadcast the image-level advantage $A_g$ to all tokens, assigning equal credit to structural and refinement tokens despite their fundamentally different roles.
We instead assign token-level credit weights according to generation order:
\begin{equation}
\tilde{A}_{g,i} = A_g \cdot w_{g,i}, \qquad
w_{g,i} =
\begin{cases}
\lambda_{\mathrm{s}}, & i \in \mathcal{S}_{g,\mathrm{structure}}, \\
\lambda_{\mathrm{r}}, & i \in \mathcal{S}_{g,\mathrm{refinement}},
\end{cases}
\end{equation}
where $\lambda_{\mathrm{s}} > 1 > \lambda_{\mathrm{r}} \ge 0$.
Structural tokens receive weights above the uniform baseline to amplify updates on global composition and layout, while refinement tokens receive smaller weights to attenuate updates on local detail.
All three training stages share the same weighting scheme. The only difference across stages is which active token set $\mathcal{S}_{g,s}$ is selected.

The HT-GRPO per-step objective is Eq.~\eqref{eq:grpo_dmlm_k} with $A_g$ replaced by $\tilde{A}_{g,i}$:
\begin{align}
\label{eq:htgrpo_obj}
\mathcal{J}^{(k)}(\theta)
= \mathbb{E}_{c,\,\{\mathbf{x}^{(0)}_g\}\sim\pi_{\theta_{\mathrm{old}}}}\!\Bigg[
  &\frac{1}{G}\sum_{g=1}^{G}
  \frac{1}{|\mathcal{M}_g^{(k)}|}\!\sum_{i\in\mathcal{M}_g^{(k)}}
  \min\!\Big(r_{g,i}(\theta)\,\tilde{A}_{g,i},\;
  \mathrm{clip}\!\left(r_{g,i}(\theta),1{-}\epsilon,1{+}\epsilon\right)\tilde{A}_{g,i}\Big) \notag\\
  &-\;\beta\,\mathbb{D}_{\mathrm{KL}}\!\left(\pi_\theta\,\|\,\pi_{\mathrm{ref}}\right)
\Bigg],
\end{align}
where the $K$ updates are ordered as Global $\to$ Structure $\to$ Refinement, consistent with Appendix~\ref{app:unified_framework}.

%% file: sec/4_experiments.tex
\input{sec/tables/tab_geneval_main}

\section{Experiments}
\label{sec:experiments}

\subsection{Experimental Setup}
\noindent\textbf{Models.} 
We validate our proposed HT-GRPO on two popular open-source pre-trained dMLLMs: MMaDA~\cite{mmada} and Lumina-DiMOO~\cite{luminadimoo}. 
Since other notable dMLLMs, such as LaViDa-O~\cite{li2025lavida}, are not open-source, we cannot include them in our experiments.
Additionally, models like LLaDA-o~\cite{you2026lladao} are not strictly dMLLMs because they rely on continuous diffusion for image generation, making them unsuitable as a base model.

\noindent\textbf{Baselines.} 
Our primary baselines include specialized T2I models (e.g., SDXL~\cite{podell2023sdxl}, FLUX.1-dev~\cite{flux}, DALL-E 3~\cite{betker2023dalle3}, Janus-Pro~\cite{janusPro}), as well as the base versions of MMaDA and Lumina-DiMOO.
Furthermore, we compare various RL training strategies.
Specifically, we evaluate MaskGRPO~\cite{maskgrpo} on the same dMLLM backbones to highlight the advantages of HT-GRPO.
We additionally compare against TraceGRPO, a trajectory-aware baseline that constructs causally consistent contexts from the recorded unmasking order. Following the implementation of TraceRL~\cite{tracerl}, we refer to this method as TraceGRPO.
We also test UniGRPO to establish the performance bound for single-stage RL.

\noindent\textbf{Benchmarks \& Metrics.}
To evaluate T2I generation performance, we employ two related benchmarks: GenEval~\cite{geneval} and DPG-Bench~\cite{dpgbench}.
GenEval, consisting of 553 text prompts, is the most widely used benchmark to assess object-centric generation.
In contrast, DPG-Bench includes 1,065 dense and complex prompts to evaluate how well the generated images align with the text.
Beyond benchmark performance, we also measure the human preference, visual quality, and aesthetics of the generated images using  ImageReward~\cite{xu2023imagereward}, DeQA~\cite{deqa_score}, HPSv3~\cite{hpsv3}, and UniPercept~\cite{UniPercept}.

\noindent\textbf{Implementation Details.} 
HT-GRPO is implemented on top of the MaskGRPO~\cite{maskgrpo} codebase, inheriting its optimizer, reward, and RL loop. 
Specifically, we generate $G=9$ rollouts per prompt.
The final reward score is a combination of HPSv3~\cite{hpsv3}, CLIP score~\cite{taited2023CLIPScore}, and UniReward~\cite{unifiedreward}, maintaining the original KL penalty coefficient and fixing the classifier-free guidance at 3.5.
HT-GRPO-specific settings are: structure ratio $\alpha=0.3$, three-stage budget $n_{\mathrm{global}}:n_{\mathrm{structure}}:n_{\mathrm{refinement}}=2:4:2$ ($K=8$), token-level credit weights $\lambda_{\mathrm{s}}=1.5$ and $\lambda_{\mathrm{r}}=0.5$, and linear-decay annealing ($\gamma_{\max}=1.0$, $\gamma_{\min}=0.5$, down mode). 
All experiments run on 8$\times$ A100-80G GPUs.
More details in Appendix~\ref{appendix:settings}.

\input{sec/tables/tab_dpg_main}

\input{sec/tables/tab_dpg_auto}
\subsection{Main Results}
% \vspace{-0.08in}
\noindent\textbf{GenEval.}
% All GenEval scores are reported as percentages (\%). Table~\ref{tab:geneval-main} shows that HT-GRPO raises the GenEval overall score from \textcolor{green}{80.00} (MaskGRPO) to \textcolor{green}{82.41} on the same MMaDA backbone, surpassing all discrete-generation references including Janus-Pro (\textcolor{green}{80.00}), MMaDA + UniGRPO (\textcolor{green}{63.00}), and Show-o (\textcolor{green}{53.00}), and matching FLUX.1-dev (\textcolor{green}{82.00})—a continuous-generation model trained at far greater scale. The gain is not uniform across dimensions: Two Objects rises by $\approx$\textcolor{green}{7 points} (\textcolor{green}{85.00$\to$91.92}) and Counting by $\approx$\textcolor{green}{10 points} (\textcolor{green}{66.00$\to$76.25}). These are precisely the dimensions that require resolving inter-object structure and spatial relations—decisions that structural tokens must commit to during the early, high-entropy phase of generation. Single-object accuracy reaches a perfect \textcolor{green}{100.00}. Color, Position, and Attribute decrease marginally (\textcolor{green}{1--3 points}), consistent with the design intent: prioritizing compositional structure involves reallocating some credit from texture-level refinement to coarse layout. On Lumina-DiMOO, the same hierarchical training strategy raises the overall score from \textcolor{green}{83.00} to \textcolor{green}{91.00} with gains across all six dimensions, establishing a new state of the art among all evaluated models and confirming cross-architecture transferability.
As shown in Table~\ref{tab:geneval}, the evaluation on the GenEval benchmark highlights the advantages of HT-GRPO in three main aspects.
First, HT-GRPO is highly effective and outperforms previous RL methods.
For the MMaDA, our method increases the base overall score from 0.56 to 0.83, clearly outperforming UniGRPO (0.63), MaskGRPO (0.80), and TraceGRPO (0.79).
Second, this performance gain is consistent across different dMLLMs.
When applied to the stronger Lumina-DiMOO, HT-GRPO improves the overall score from 0.83 to 0.92 and outperforms MaskGRPO (0.88) and TraceGRPO (0.87).
This proves that our method generalizes well to different dMLLMs.
Finally, models trained with HT-GRPO are fully comparable to state-of-the-art T2I models.
Leading models like FLUX.1-dev achieve an overall score of 0.82.
MMaDA with HT-GRPO reaches 0.83 to slightly pass them, and Lumina-DiMOO with HT-GRPO sets a new high score of 0.92.

\noindent\textbf{DPG-Bench.}
DPG-Bench is more challenging than GenEval as it evaluates models using dense and complex prompts.
%
% As shown in Table~\ref{tab:dpg_final_results}, MaskGRPO provides only a minimal improvement over the strong Lumina-DiMOO baseline, with the score rising slightly from 81.86 to 82.24.
% 
As shown in Table~\ref{tab:dpg_final_results}, both MaskGRPO and TraceGRPO yield only marginal improvements over the Lumina-DiMOO baseline (82.24 and 82.62), with TraceGRPO even underperforming MaskGRPO on MMaDA (74.37 vs. 75.81).
In contrast, HT-GRPO successfully overcomes this performance bottleneck, raising the score to 84.47.
This result surpasses leading models such as DALL-E 3 (83.50) and FLUX.1-dev (83.84), with similar gains observed when using the MMaDA base model.
Furthermore, Table~\ref{tab:dpg_quality} highlights another key advantage: HT-GRPO improves semantic alignment without sacrificing visual quality.
It consistently achieves almost the highest scores across human preference metrics (e.g., ImageReward, HPSv3) and image quality \& aesthetic evaluations (e.g., DeQA, UniPercept), proving that HT-GRPO effectively balances complex instruction following with high-fidelity image generation.

\subsection{Ablation Study}
% \vspace{-2mm}
Figure~\ref{fig:ablation} and Table~\ref{tab:ablate-a3} (shown in Appendix~\ref{app:ablation}) summarize five ablation studies on MMaDA ($K=8$, $\alpha=0.3$).
\textbf{(a) Stage organization and budget:} The full Global$\to$Structure$\to$Refinement schedule with a 2:4:2 budget reaches 83.3 overall, outperforming all single-stage and two-stage variants (78.3--80.2); the coarse-to-fine ordering itself is critical.
\textbf{(b) Structure ratio:} Performance peaks at $\alpha=0.3$ and degrades for narrower ($\alpha=0.1$, 80.2) and broader ($\alpha=0.5$) boundaries, confirming that a well-calibrated hierarchy is essential.
\textbf{(c) Component analysis:} Removing hierarchical credit weighting ($\lambda_{\mathrm{s}}=\lambda_{\mathrm{r}}=1$) drops the score to 80.8, and replacing $\mathbf{C}_\emptyset$ with revealed structural contexts reduces it to 80.6, validating both designs independently.
The linear-decay annealing schedule (Table~\ref{tab:ablate-a3}, 83.31) outperforms static and ascending baselines, confirming the value of broad initial coverage for stable gradient directions.
Full per-ablation details are provided in Appendix~\ref{app:ablation}.

\begin{figure}[t]
    \centering
    % \vspace{-0.1in}
    \includegraphics[width=1\linewidth]{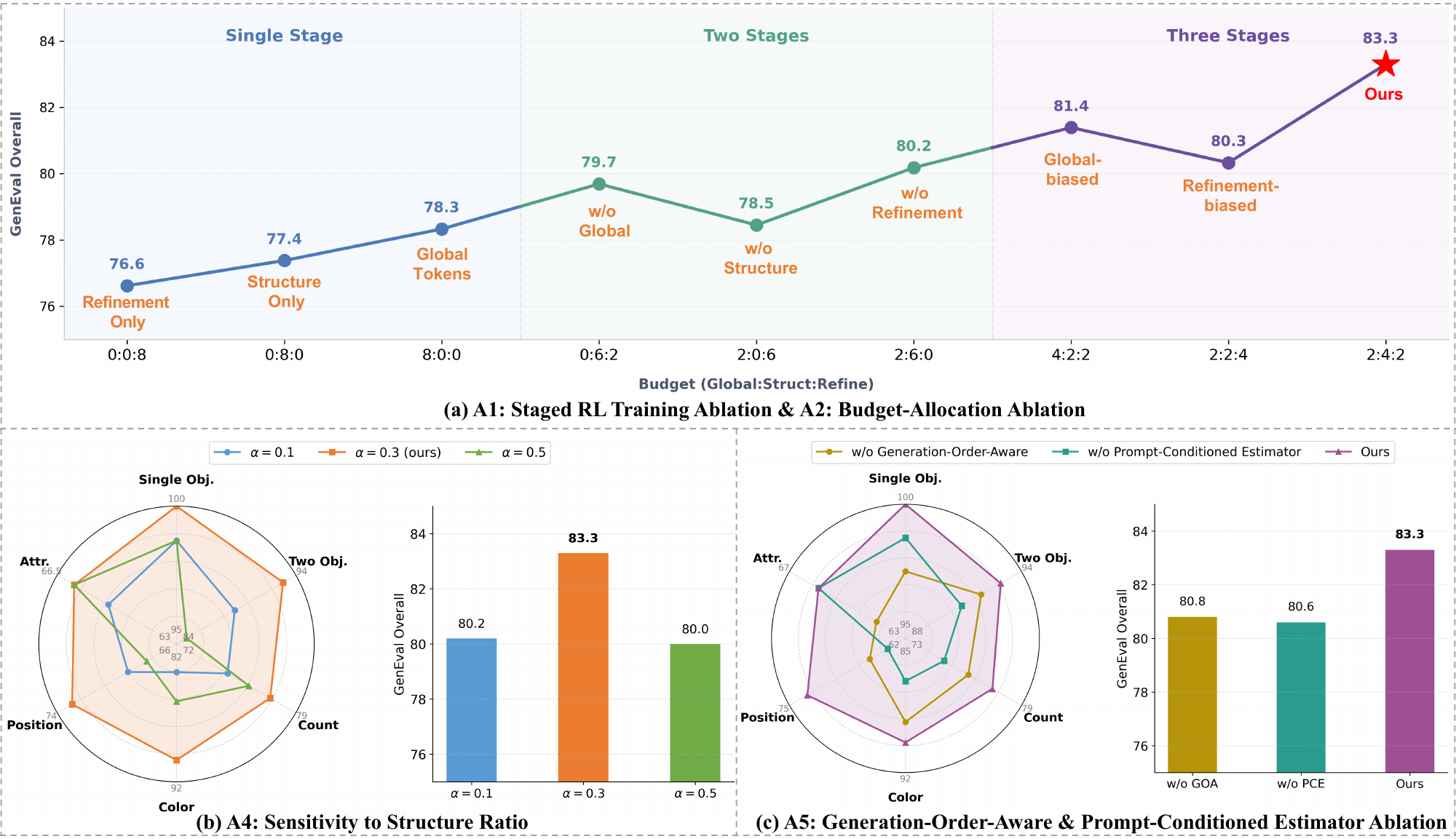}
    % \vspace{-0.22in}
      \caption{\textbf{(a) Stage organization and budget allocation:} The full Global$\to$Structure$\to$Refinement schedule with a structure-biased 2:4:2 budget consistently outperforms single-stage and two-stage variants. \textbf{(b) Sensitivity to structure ratio $\alpha$:} Performance peaks at $\alpha = 0.3$ and degrades when the boundary is too narrow or too broad. \textbf{(c) Component analysis:} Hierarchical Credit Assignment and Prompt-Conditioned Estimator each provide independent gains, validating both design choices.}
    \label{fig:ablation}
    % \vspace{-0.15in}
\end{figure}

%% file: sec/tables/tab_geneval_main.tex
\begin{table*}[!t]
% \vspace{-8mm}
\centering
\renewcommand{\arraystretch}{1.1}
\setlength{\tabcolsep}{2mm} 
\vspace{-0.1in}
\caption{\textbf{Performance comparison on GenEval} across various dMLLMs and RL settings.}
\vspace{1mm}
\resizebox{1.01\textwidth}{!}{
\begin{tabular}{c | c  c c c c c | c}
\toprule
 \textbf{Methods}  & \textbf{Single $\uparrow$} & \textbf{Two $\uparrow$} & \textbf{Counting $\uparrow$} & \textbf{Colors $\uparrow$} & \textbf{Position $\uparrow$} & \textbf{Attribute $\uparrow$} & \textbf{Overall $\uparrow$} \\
\cmidrule{1-8}
\rowcolor{blue!10}
\multicolumn{8}{c}{\textbf{\textit{Text-to-Image Models}}} \\
\cmidrule{1-8}
SDXL~\cite{podell2023sdxl}  &  0.98 & 0.74 &  0.39 &  0.85 &  0.15 &  0.23 & 0.55 \\
DALL-E 3~\cite{betker2023dalle3} &  0.96 & 0.87 &  0.47 &  0.83 &  0.43 &  0.45 & 0.67 \\
% SANA-1.5~\cite{xie2025sana}  &  0.99 & 0.85 &  0.77 & 0.87  & 0.34  &  0.54  & 0.72 \\
Janus-Pro~\cite{janusPro}  &  0.99  & 0.89 & 0.59  & 0.90  &  0.79 & 0.66 & 0.80 \\
FLUX.1-dev~\cite{flux}  & 0.98 & 0.93 & 0.75 & \textbf{0.93} & 0.68 & 0.65 & 0.82 \\
% Z-Image~\cite{cai2025z} & \textbf{1.00} & \textbf{0.95} & 0.77 & 0.89 & 0.65 & 0.68 & 0.82\\
\cmidrule{1-8}
\rowcolor{green!10} 
\multicolumn{8}{c}{\textbf{\textit{dMLLMs (wo / w RL)}}} \\
\cmidrule{1-8}

MMaDA~\cite{mmada} & 0.96 & 0.60 & 0.45 & 0.81 & 0.14 & 0.25 & 0.56  \\
+ UniGRPO~\cite{mmada} & 0.99 & 0.76 & 0.61 & 0.84 & 0.20 & 0.37 & 0.63\\ 
+ MaskGRPO~\cite{maskgrpo} & 0.99 & 0.85 & 0.66 & 0.89 & \textbf{0.73} & \textbf{0.69} & 0.80\\
+ TraceGRPO~\cite{tracerl}
& 1.00 & 0.87 & 0.72 & \textbf{0.92} & 0.64 & 0.62 & 0.79 \\
\rowcolor{gray!15} 
+ HT-GRPO (Ours) & \textbf{1.00} & \textbf{0.92} & \textbf{0.77} & 0.90 & \textbf{0.73} & 0.66 & \textbf{0.83} \\

\cmidrule{1-8}
Lumina-DiMOO~\cite{luminadimoo} & 0.99 & 0.94 & 0.70 & 0.85 & 0.81 & 0.71 & 0.83 \\
+ MaskGRPO~\cite{maskgrpo} & \textbf{1.00} & 0.90 & 0.88 & 0.89 & 0.85 & \textbf{0.77} & 0.88 \\
+ TraceGRPO~\cite{tracerl}
& \textbf{1.00} & 0.90 & 0.85 & 0.90 & 0.88 & 0.73 & 0.87 \\

\rowcolor{gray!15}
+ HT-GRPO (Ours) & \textbf{1.00} & \textbf{0.95} & \textbf{0.94} & \textbf{0.93} & \textbf{0.93} & 0.75 & \textbf{0.92}\\
\bottomrule
\end{tabular}
}
\label{tab:geneval}
\vspace{-0.15in}
\end{table*}

%% file: sec/tables/tab_dpg_main.tex
% \begin{table*}[t]
% \centering
% \small
% \caption{Main results on DPG-Bench official metrics.}
% \label{tab:dpg-main}
% \resizebox{\linewidth}{!}{
% \begin{tabular}{lcccccc}
% \toprule
% Model & Global$\uparrow$ & Entity$\uparrow$ & Attribute$\uparrow$ & Relation$\uparrow$ & Other$\uparrow$ & Overall$\uparrow$ \\
% \midrule
% \multicolumn{7}{l}{Public reference models} \\
% SDXL (2023) & 83.27 & 82.43 & 80.91 & 86.76 & 80.41 & 74.65 \\
% Emu3-Gen (2024) & 85.21 & 86.68 & 86.84 & 90.22 & 83.15 & 80.60 \\
% \midrule
% \multicolumn{7}{l}{MMaDA backbone and its baselines} \\
% MMaDA (2025) & 77.52 & 77.52 & 77.67 & 81.56 & 63.13 & 70.51 \\
% MMaDA + MaskGRPO & 80.92 & 79.52 & 85.41 & 83.05 & 69.12 & 75.81 \\
% MMaDA + SFT+MaskGRPO & 85.96 & 84.05 & \textbf{89.76} & 83.43 & 79.88 & 81.76 \\
% \midrule
% \multicolumn{7}{l}{Lumina-DiMOO backbone and its baselines} \\
% Lumina-DiMOO (base) & 81.17 & 89.96 & 88.10 & 89.64 & 79.95 & 81.86 \\
% \midrule
% \multicolumn{7}{l}{HT-GRPO (ours)} \\
% MMaDA + HT-GRPO & \textbf{89.76} & 87.62 & 86.58 & 88.86 & 81.83 & 81.09 \\
% Lumina-DiMOO + HT-GRPO & 82.65 & \textbf{91.22} & 88.33 & \textbf{92.56} & \textbf{82.05} & \textbf{83.71} \\
% \bottomrule
% \end{tabular}
% }
% \end{table*}

\begin{table*}[!t]
% \vspace{-8mm}
\centering
\renewcommand{\arraystretch}{1.1}
\setlength{\tabcolsep}{3mm} 
% \vspace{-0.1in}
\caption{\textbf{Performance comparison on DPG-Bench} across various dMLLMs and RL settings.}
% \vspace{1mm}
\resizebox{1.01\textwidth}{!}{
\begin{tabular}{c | c  c c c c  | c}
\toprule
\textbf{Model} & \textbf{Global $\uparrow$} & \textbf{Entity $\uparrow$} & \textbf{Attribute $\uparrow$} & \textbf{Relation $\uparrow$} & \textbf{Other $\uparrow$} & \textbf{Overall $\uparrow$} \\
\cmidrule{1-7}
\rowcolor{blue!10}
\multicolumn{7}{c}{\textbf{\textit{Text-to-Image Models}}} \\
\cmidrule{1-7}
SDXL~\cite{podell2023sdxl} & 83.27 & 82.43 & 80.91 & 86.76 & 80.41 & 74.65 \\
DALL-E 3~\cite{betker2023dalle3} &90.97 &89.61 &88.39 &90.58 &89.83 &83.50\\
Janus-Pro~\cite{janusPro} &86.90 &88.90 &89.40 &89.32 &89.48 &84.19 \\
FLUX.1-dev~\cite{flux} &74.35 &90.00 &88.96 &90.87 &88.33 &83.84 \\
% Z-Image~\cite{cai2025z} &\textbf{91.29} &89.59 &90.14 &92.16 &88.68 &84.86\\
\cmidrule{1-7}
\rowcolor{green!10} 
\multicolumn{7}{c}{\textbf{\textit{dMLLMs (wo / w RL)}}} \\
\cmidrule{1-7}

MMaDA~\cite{mmada} & 77.52 & 77.52 & 77.67 & 81.56 & 63.13 & 70.51 \\
% + UniGRPO~\cite{mmada} \\ 
+ MaskGRPO~\cite{maskgrpo} & 80.92 & 79.52 & 85.41 & 83.05 & 69.12 & 75.81 \\
% + SFT \& MaskGRPO & 85.96 & 84.05 & \textbf{89.76} & 83.43 & 79.88 & 81.76 \\
+ TraceGRPO~\cite{tracerl}
& 82.34 & 78.92 & 81.29 & 82.08 & 70.21 & 74.37 \\
\rowcolor{gray!15} 
+ HT-GRPO (Ours) & \textbf{89.76} & \textbf{87.62} & \textbf{86.58} & \textbf{88.86} & \textbf{81.83} & \textbf{81.09} \\

\cmidrule{1-7}
Lumina-DiMOO~\cite{luminadimoo} & 81.17 & 89.96 & 88.10 & 89.64 & 79.95 & 81.86 \\
+ MaskGRPO~\cite{maskgrpo} & 82.65 & 89.50 & 88.33 & 88.60 & \textbf{88.24} & 82.24 \\
+ TraceGRPO~\cite{tracerl}
& 84.89 & 89.65 & 88.23 & 90.24 & 84.28 & 82.62 \\

\rowcolor{gray!15}
+ HT-GRPO (Ours) & \textbf{88.04}  & \textbf{91.22} & \textbf{88.98} & \textbf{92.56} & 83.05 & \textbf{84.47}\\
\bottomrule
\end{tabular}
}
\label{tab:dpg_final_results}
% \vspace{-0.18in}
\end{table*}

%% file: sec/tables/tab_dpg_auto.tex
% \begin{table*}[t]
% \centering
% \small
% \caption{Automatic metrics on DPG-Bench samples.}
% \label{tab:dpg-auto}
% \resizebox{\linewidth}{!}{
% \begin{tabular}{llccc}
% \toprule
% Backbone & Method & DeQA$\uparrow$ & ImageReward$\uparrow$ & HPSv3$\uparrow$ \\
% \midrule
% \multicolumn{5}{l}{MMaDA backbone and its baselines} \\
% MMaDA & Base model & 3.9107 & 0.6408 & 8.8100 \\
% MMaDA & + MaskGRPO & 4.0499 & 0.8317 & 9.4000 \\
% \midrule
% \multicolumn{5}{l}{Lumina-DiMOO backbone and its baselines} \\
% Lumina-DiMOO & Base model & 4.0289 & 0.8428 & 11.6013 \\
% \midrule
% \multicolumn{5}{l}{HT-GRPO (ours)} \\
% MMaDA & + HT-GRPO & 4.0521 & \textbf{0.8586} & \textbf{10.4550} \\
% Lumina-DiMOO & + HT-GRPO & \textbf{4.1150} & \textbf{0.9742} & \textbf{12.4588} \\
% \bottomrule
% \end{tabular}
% }
% \end{table*}

\begin{table*}[!t]
\centering
\renewcommand{\arraystretch}{1.1}
\setlength{\tabcolsep}{1.5mm} 
\caption{\textbf{Human preference, image quality, and aesthetics comparison on DPG-Bench} across various dMLLMs and RL settings.}
% \vspace{1mm}
\resizebox{1.01\textwidth}{!}{
\begin{tabular}{c | c c c c c c}
\toprule
\multirow{2}{*}{\textbf{Methods}} 
& \multirow{2}{*}{\textbf{DeQA $\uparrow$}} 
& \multirow{2}{*}{\textbf{ImgReward $\uparrow$}} 
& \multirow{2}{*}{\textbf{HPSv3 $\uparrow$}} 
& \multicolumn{3}{c}{\textbf{UniPercept}} \\
\cmidrule(lr){5-7}
& & & & \textbf{Aesthetics $\uparrow$} & \textbf{Quality $\uparrow$} & \textbf{Structure $\uparrow$} \\
\midrule

MMaDA~\cite{mmada}  
& 3.91 & 0.64 & 8.81 & 58.83 & 65.29 & 47.46 \\

+ MaskGRPO~\cite{maskgrpo} 
& 4.05 & 0.83 & 9.40 & 59.02 & 67.07 & 48.15 \\

+ TraceGRPO~\cite{tracerl}
& \textbf{4.08} & 0.84 & 9.56 & 58.78 & 66.87 & 49.04 \\

\rowcolor{gray!15} 
+ HT-GRPO (Ours) 
& 4.07 & \textbf{0.86} & \textbf{10.46} & \textbf{59.29} & \textbf{67.91} & \textbf{49.45} \\

\midrule

Lumina-DiMOO~\cite{luminadimoo} 
& 4.03 & 0.84 & 11.60 & 56.66 & 66.27 & 43.12 \\

+ MaskGRPO~\cite{maskgrpo} 
& 4.08 & 0.90 & 12.00  & 56.77 & 66.13 & 43.55 \\

+ TraceGRPO~\cite{tracerl}
& 4.10 & 0.88 & 11.28 & 57.68 & 67.21 & 43.89 \\

\rowcolor{gray!15}
+ HT-GRPO (Ours) 
& \textbf{4.12} & \textbf{0.97} & \textbf{12.46} & \textbf{58.39} & \textbf{67.97} & \textbf{44.97} \\

\bottomrule
\end{tabular}
}
\label{tab:dpg_quality}
% \vspace{-0.2in}
\end{table*}

%% file: sec/6_conclusion.tex
\vspace{-2mm}
\section{Conclusion}
\label{sec:conclusion}
\vspace{-2mm}
We revisit the application of GRPO to dMLLMs and identify two key challenges: multiple valid unmasking orderings make importance-ratio estimation difficult, while existing methods uniformly assign image-level rewards to tokens with heterogeneous roles. 
To address these issues, we propose HT-GRPO, which uses generation-order rank to characterize token uncertainty and role, organizes inner-loop updates into Global, Structure, and Refinement stages, estimates importance ratios under a fully masked prompt-conditioned context, and performs generation-order-aware credit assignment. 
Experiments on MMaDA and Lumina-DiMOO show consistent improvements across GenEval, DPG-Bench, and six human preference and visual quality metrics, demonstrating the effectiveness of hierarchical token optimization. 
A current limitation is that HT-GRPO uses a fixed structure ratio and fixed credit weights, which may not optimally adapt to prompts with different compositional complexity or rollout uncertainty.
In the future, we will advance HT-GRPO through the integration of adaptive token grouping and dynamic credit assignment.

%% file: sec/A_supplementary.tex
\clearpage
\onecolumn
% \setcounter{page}{1}
% \maketitlesupplementary
\setcounter{section}{0}
\renewcommand{\thesection}{\Alph{section}}

 \begin{center}
    \Large \textbf{Supplementary Material} \\ % 自定义标题格式
    \vspace{1.0em}
\end{center}

\input{sec/5_related_work}
\section{A Unified Theoretical Framework for dMLLM RL Methods}
\label{app:unified_framework}
This section formalizes the two existing approaches described in Section~\ref{sec:preliminaries}, random remasking methods and trajectory recording methods, together with HT-GRPO. We use $g$, $i$, $k$, and $c$ for rollout index, token position, inner-loop update index, and text prompt condition respectively. All other symbols follow the notation in the main text.

As established in Section~\ref{subsec:limitations}, the surrogate ideal objective $\mathcal{J}^*$ (Eq.~\eqref{eq:ideal}) requires averaging over all valid denoising trajectories $\tau \in \mathcal{T}$, which is combinatorially intractable. All three paradigms therefore share the same clipped surrogate objective and differ only in three design choices: the conditioning context $\mathbf{C}_{g,i}^{(k)}$, the optimization support set $\mathcal{M}_g^{(k)}$, and the token-level weight $w_{g,i}$. Setting these to the choices in Section~\ref{sec:method} recovers HT-GRPO exactly.
\begin{equation}
\label{eq:unified_obj}
\mathcal{L}^{(k)}(\theta)
=
\frac{1}{G}\sum_{g=1}^{G}
\frac{1}{|\mathcal{M}_g^{(k)}|}
\sum_{i\in \mathcal{M}_g^{(k)}}
\min\!\left(
r_{g,i}^{(k)}(\theta)\,\tilde A_{g,i},\;
\mathrm{clip}\!\left(r_{g,i}^{(k)}(\theta),1-\epsilon,1+\epsilon\right)\tilde A_{g,i}
\right),
\end{equation}
where
\begin{equation}
r_{g,i}^{(k)}(\theta)=
\frac{
\pi_\theta\!\left(v_{g,i}\mid \mathbf{C}_{g,i}^{(k)},c\right)
}{
\pi_{\theta_{\mathrm{old}}}\!\left(v_{g,i}\mid \mathbf{C}_{g,i}^{(k)},c\right)
}.
\end{equation}
Here, $k\in\{1,\ldots,K\}$ indexes the inner-loop gradient step, and the same rollout batch is reused for all $K$ updates. $v_{g,i}$ is the token value generated at position $i$ in rollout $g$. $\mathcal{M}_g^{(k)}$ is the optimization support set, the set of token positions that contribute gradients in the $k$-th inner-loop update. The conditioning context $\mathbf{C}_{g,i}^{(k)}$ is the partially observed token configuration used to evaluate the importance ratio for position $i$ in update $k$. The token-level weighted advantage $\tilde{A}_{g,i} = w_{g,i} A_g$ scales the group-relative advantage $A_g$ by a per-token credit weight $w_{g,i}$, where
\begin{equation}
A_g = \frac{R_g - \mathrm{mean}(\{R_j\})}{\mathrm{std}(\{R_j\})+\delta_m}.
\end{equation}
The three methods differ only in how they instantiate these three quantities.

\subsection{Random Remasking Methods}
\label{app:random_remasking}

Random remasking methods, including MaskGRPO~\cite{maskgrpo}, D1~\cite{d1}, and UniGRPO~\cite{unigrpo}, retain only the final image $\mathbf{x}_g$ and construct synthetic contexts by independently remasking each position with probability $\gamma_k$:
\begin{equation}
\begin{aligned}
\mathcal{M}_g^{(k)} &= \bigl\{i : i \text{ is remasked with prob.}\ \gamma_k\bigr\}
  && \text{(support set)}, \\
\mathbf{C}_{g,i}^{(k)} &= \mathrm{Mask}(\mathbf{x}_g,\mathcal{M}_g^{(k)})
  && \text{(mask $\mathcal{M}_g^{(k)}$; retain token values elsewhere)}, \\
w_{g,i} &= 1
  && \text{(uniform weight)}.
\end{aligned}
\end{equation}

\noindent\textit{Stage conflation.}\quad
Because remasking is independent of the generation-order rank $\rho_{g,i}$, the support set $\mathcal{M}_g^{(k)}$ mixes structural tokens (small $\rho_{g,i}$) and refinement tokens (large $\rho_{g,i}$) in the same update with equal probability. By \hyperref[prop:entropy-monotonicity]{Proposition~C.1}, these two groups carry fundamentally different levels of uncertainty, so their gradients operate at very different scales within a single update step.

\noindent\textit{Future-token contamination.}\quad
The context $\mathbf{C}_{g,i}^{(k)}$ retains all positions not in $\mathcal{M}_g^{(k)}$. For position $i$, define the future token set as $\mathcal{F}_i^{(g)}=\{j:\rho_{g,j}>\rho_{g,i}\}$. Since remasking ignores generation order, each position in $\mathcal{F}_i^{(g)}$ is retained in $\mathbf{C}_{g,i}^{(k)}$ with probability $1-\gamma_k$, potentially exposing tokens that had not yet been generated when position $i$ was predicted. \hyperref[prop:contamination]{Proposition~C.2} shows this occurs with strictly positive probability for any non-final position.

\noindent\textit{Multi-path coverage (partial).}\quad
Different random masks induce diverse conditioning contexts, exploring a broader range of token-context combinations than a single fixed trajectory. Random remasking is agnostic to generation order, however, so the resulting contexts are not guaranteed to correspond to any valid denoising trajectory. Because the context may include future tokens, as shown in the contamination analysis above, the diverse paths explored do not respect the causal ordering of the original generation process. The approximation of $\mathcal{J}^*$ is therefore partial: broader in context diversity than a single trajectory, but biased away from the causal structure that $\mathcal{J}^*$ requires.

\noindent\textit{Uniform token reward (unresolved).}\quad
All positions receive the same credit weight $w_{g,i}=1$ regardless of their generation-order rank. Structural and refinement tokens carry fundamentally different entropy levels and play different roles in image generation, as established in Section~\ref{subsec:limitations}. Equal credit assignment ignores this asymmetry entirely.

\subsection{Trajectory Recording Methods}
\label{app:trajectory_recording}
Trajectory recording methods, including TraceRL~\cite{tracerl}, CJ-GRPO~\cite{cjgrpo}, d-TreeRPO~\cite{dtreerpro}, and AGRPO~\cite{agrpo}, record the denoising order and evaluate each token under its trajectory-consistent conditioning context:
\begin{equation}
\begin{aligned}
\mathcal{M}_g^{(k)} &= \{1,\ldots,N\}
  && \text{(all $N$ tokens per update)}, \\
\mathbf{C}_{g,i}^{(k)} &= \mathbf{x}_{g}^{(\prec i)}
  && \text{(reveals $\{j:\rho_{g,j}<\rho_{g,i}\}$; remaining positions masked)}, \\
w_{g,i} &= 1
  && \text{(uniform weight)}.
\end{aligned}
\end{equation}

\noindent\textit{Stage conflation (avoided).}\quad
Unlike random remasking methods, which apply the same synthetic context to all tokens in $\mathcal{M}_g^{(k)}$ regardless of generation rank, trajectory recording methods set $\mathcal{M}_g^{(k)} = \{1,\ldots,N\}$ and assign each token its own trajectory-consistent context $\mathbf{C}_{g,i}^{(k)} = \mathbf{x}_g^{(\prec i)}$. Structural tokens therefore receive sparse contexts reflecting their early generation stage, while refinement tokens receive progressively richer contexts. Each importance ratio is computed under a context that matches the entropy level of the corresponding token, avoiding the uniform-context mismatch that characterizes stage conflation.

\noindent\textit{Future-token contamination (avoided).}\quad
The trajectory-consistent context $\mathbf{C}_{g,i}^{(k)} = \mathbf{x}_g^{(\prec i)}$ reveals only positions with $\rho_{g,j} < \rho_{g,i}$. All positions in the future token set $\mathcal{F}_i^{(g)} = \{j : \rho_{g,j} > \rho_{g,i}\}$ are masked by construction, so future-token contamination cannot occur.

\noindent\textit{Limited path coverage.}\quad
Both $\mathcal{M}_g^{(k)}$ and $\mathbf{C}_{g,i}^{(k)}$ are derived from a single rollout trajectory, so each update evaluates only one element of the expectation in Eq.~\eqref{eq:ideal}. The many other valid trajectories $\tau \in \mathcal{T}$ that could produce the same image are never seen, restricting the diversity of generation paths the model learns from.

\noindent\textit{Uniform token reward (unresolved).}\quad
All positions receive the same credit weight $w_{g,i}=1$. Although the trajectory-consistent context $\mathbf{x}_g^{(\prec i)}$ naturally differentiates the conditioning richness of structural versus refinement tokens, the advantage broadcast to each token remains uniform. There is no explicit mechanism to amplify the learning signal for high-uncertainty structural tokens or attenuate it for near-deterministic refinement tokens.

\subsection{HT-GRPO}
HT-GRPO retains the final sample together with the generation-order rank $\rho_{g,i}\in\{1,\ldots,N\}$, which records the unmasking step at which position $i$ was revealed in rollout $g$, with $\rho_{g,i}=1$ for the first token and $\rho_{g,i}=N$ for the last. Positions unmasked in the same denoising step share the same context and entropy level despite receiving different rank values. HT-GRPO partitions tokens into global, structural, and refinement groups:
\begin{equation}
\begin{aligned}
\mathcal{S}_{g,\mathrm{global}} &= \{1,\ldots,N\}, \\
\mathcal{S}_{g,\mathrm{structure}} &= \{i:\rho_{g,i}\le N_s\}, \\
\mathcal{S}_{g,\mathrm{refinement}} &= \{i:\rho_{g,i}>N_s\},
\end{aligned}
\end{equation}
where $N_s=\lfloor \alpha N\rfloor$ and $K = n_{\mathrm{global}}+n_{\mathrm{structure}}+n_{\mathrm{refinement}}$ is the total number of inner-loop updates. The $K$ steps follow a fixed Global $\to$ Structure $\to$ Refinement schedule, and the support set at step $k$ is a random subset drawn from the active stage:
\begin{equation}
\mathcal{M}_g^{(k)} \subseteq
\begin{cases}
\mathcal{S}_{g,\mathrm{global}}, & 1 \le k \le n_{\mathrm{global}}, \\
\mathcal{S}_{g,\mathrm{structure}}, & n_{\mathrm{global}} < k \le n_{\mathrm{global}}+n_{\mathrm{structure}}, \\
\mathcal{S}_{g,\mathrm{refinement}}, & n_{\mathrm{global}}+n_{\mathrm{structure}} < k \le K.
\end{cases}
\end{equation}
All tokens share the same fully masked conditioning context, defined as the state in which all $N$ image positions remain masked:
\begin{equation}
\mathbf{C}_{g,i}^{(k)} \equiv \mathbf{C}_{\emptyset}.
\end{equation}
Token weights are generation-order-aware rather than uniform:
\begin{equation}
\begin{aligned}
\tilde A_{g,i} &= w_{g,i}A_g, \\
w_{g,i} &=
\begin{cases}
\lambda_{\mathrm{s}}, & i\in\mathcal{S}_{g,\mathrm{structure}}, \\
\lambda_{\mathrm{r}}, & i\in\mathcal{S}_{g,\mathrm{refinement}}.
\end{cases}
\end{aligned}
\end{equation}

\noindent\textit{Stage conflation (resolved).}\quad
At each inner-loop step $k$, the support set $\mathcal{M}_g^{(k)}$ is drawn exclusively from the token set of the active stage. The Global $\to$ Structure $\to$ Refinement schedule partitions the $K$ updates by generation rank, so high-entropy structural tokens and low-entropy refinement tokens are never placed in the same $\mathcal{M}_g^{(k)}$. The entropy gap between these two groups is established in \hyperref[prop:entropy-monotonicity]{Proposition~C.1}. This resolves the stage conflation identified in Section~\ref{subsec:limitations}.

\noindent\textit{Future-token contamination (eliminated).}\quad
Setting $\mathbf{C}_{g,i}^{(k)}\equiv\mathbf{C}_\emptyset$ masks all $N$ positions uniformly for every token and every update, making future-token exposure impossible by construction. Contamination is eliminated without requiring trajectory storage, as formalized in \hyperref[prop:contamination]{Proposition~C.2}.

\noindent\textit{Limited path coverage (alleviated).}\quad
Within each stage, $\mathcal{M}_g^{(k)}$ is re-sampled independently at every inner-loop step according to an annealed sampling rate $\gamma_{k_s}^{(s)}$ (Appendix~\ref{app:scheduling}). Different token subsets receive gradient signal across the $K$ updates, providing Monte Carlo coverage of the token space within each stage and partially alleviating the single-trajectory limitation. Because $\mathbf{C}_\emptyset$ decouples the importance ratio from any particular ordering, this intra-stage sampling introduces no causal inconsistency.

\noindent\textit{Uniform token reward (resolved).}\quad
The per-token weight $w_{g,i}$ assigns $\lambda_{\mathrm{s}}>1$ to structural tokens and $\lambda_{\mathrm{r}}<1$ to refinement tokens, replacing the uniform $w_{g,i}=1$ used by both existing families. This amplifies gradient updates on global composition and attenuates updates on near-deterministic local detail, following the rationale in Section~\ref{subsec:credit}. Trajectory recording methods avoid stage conflation and future-token contamination but leave credit assignment uniform. HT-GRPO is therefore the only paradigm that addresses all four limitations.

\subsection{Comparison}
\label{app:comparison}
Table~\ref{tab:appendix-framework} summarizes the three paradigms along eight dimensions. The formal basis for the entropy-related entries is provided in Section~\ref{app:theory} (\hyperref[prop:entropy-monotonicity]{Propositions~C.1}--\hyperref[prop:entropy-lower-bound]{C.3}).
Among the four limitations identified in Section~\ref{subsec:limitations}, random remasking methods introduce stage conflation and future-token contamination while achieving only partial path coverage. Trajectory recording methods avoid the first two but sacrifice path diversity and require $O(T)$ forward passes per rollout. Neither family addresses the uniform token reward problem. HT-GRPO resolves all four: inter-stage partitioning eliminates stage conflation, $\mathbf{C}_\emptyset$ eliminates future-token contamination, intra-stage random subsets partially alleviate limited path coverage, and generation-order-aware weights $w_{g,i}$ replace uniform credit assignment.

\input{sec/tables/tab_appendix_framework}

\section{Information-Theoretic Basis of Hierarchical Token Grouping and the Prompt-Conditioned Estimator}
\label{app:theory}
We establish three formal results underpinning HT-GRPO's design.

\paragraph{Assumption (Approximate Consistency).}
Let $V_i$ be the token at position $i$, $p_\theta$ the joint generation distribution, and $\mathbf{C}^{[n]}$ any partially revealed context with $n$ positions unmasked and position $i$ still masked. We assume the model's prediction matches the true conditional:
\begin{equation}
\pi_\theta\!\left(\cdot \mid \mathbf{C}^{[n]}, c\right)
=
p_\theta\!\left(\cdot \mid \mathbf{C}^{[n]}, c\right).
\label{eq:consistency}
\end{equation}
This connects the model's output to standard information-theoretic quantities. We write $H(\pi_\theta(\cdot\mid\mathbf{C},c))$ for the \textbf{predictive entropy}, which measures how uncertain the model is about position $i$ given visible context $\mathbf{C}$. \hyperref[prop:entropy-monotonicity]{Propositions~C.1} and~\hyperref[prop:entropy-lower-bound]{C.3} additionally require that $\mathbf{C}^{[n']}$, where $n'>n$, is obtained from $\mathbf{C}^{[n]}$ by revealing more tokens along the same trajectory while keeping position $i$ masked.

\phantomsection\label{prop:entropy-monotonicity}
\paragraph{Proposition C.1 (Entropy Monotonicity).}
\textit{Seeing more context reduces uncertainty.}
Under Assumption~\eqref{eq:consistency}, for any position $i$ and any $0 \le n < n' \le N{-}1$,
\begin{equation}
\mathbb{E}_{\mathbf{C}^{[n]}}\!\left[H\!\left(\pi_\theta(\cdot \mid \mathbf{C}^{[n]},c)\right)\right]
\ge
\mathbb{E}_{\mathbf{C}^{[n']}}\!\left[H\!\left(\pi_\theta(\cdot \mid \mathbf{C}^{[n']},c)\right)\right].
\end{equation}

\noindent\textit{Proof.}
Let $\mathbf{D}$ be the additional tokens revealed between $\mathbf{C}^{[n]}$ and $\mathbf{C}^{[n']}$. Since mutual information is non-negative,
$I(V_i;\mathbf{D}\mid\mathbf{C}^{[n]},c)=H(V_i\mid\mathbf{C}^{[n]},c)-H(V_i\mid\mathbf{C}^{[n']},c)\ge 0$.
Applying Assumption~\eqref{eq:consistency} to both sides yields the stated inequality.\hfill$\square$

\smallskip
Tokens unmasked early (small $\rho_{g,i}$, few context tokens visible) therefore carry higher uncertainty than those unmasked late. Tokens sharing the same denoising step share the same context and entropy level despite having different $\rho_{g,i}$ values. This justifies treating early tokens as structural tokens and late tokens as refinement tokens ($\mathcal{S}_{g,\mathrm{structure}}$, $\mathcal{S}_{g,\mathrm{refinement}}$ in Section~\ref{subsec:ht-grpo}).

\paragraph{Definition (Future-token set).}
For rollout $g$ and position $i$, define $\mathcal{F}_i^{(g)}=\{j:\rho_{g,j}>\rho_{g,i}\}$ as the set of tokens unmasked \emph{after} $i$. These tokens are still masked when $i$ is predicted; if any appear in the conditioning context, they inject future information into the current optimization step.

\phantomsection\label{prop:contamination}
\paragraph{Proposition C.2 (Future-Token Contamination).}
\textit{Random remasking exposes future tokens with nonzero probability, misspecifying the conditioning distribution.}
In random remasking methods, each position is independently remasked with probability $p_k\in(0,1)$. For any non-final position $i$, at least one future token from $\mathcal{F}_i^{(g)}$ remains visible in $\mathbf{C}_{g,i}^{(k)}$ with probability
\begin{equation}
\Pr\!\bigl[\mathbf{C}_{g,i}^{(k)}\not\equiv\mathbf{x}_{g}^{(\prec i)}\bigr]
=
1-p_k^{|\mathcal{F}_i^{(g)}|}>0.
\end{equation}
In the true trajectory context $\mathbf{x}_g^{(\prec i)}$ all future positions are masked; when contamination occurs, the ratio $\tilde{r}_{g,i}(\theta)=\pi_\theta(v_{g,i}\mid\mathbf{C}_{g,i}^{(k)},c)/\pi_{\theta_{\mathrm{old}}}(v_{g,i}\mid\mathbf{C}_{g,i}^{(k)},c)$ is evaluated under a causally inconsistent context, leading to a misspecified policy gradient estimator.

\noindent\textit{Proof.}
The $|\mathcal{F}_i^{(g)}|$ future positions are remasked independently; all are correctly masked with probability $p_k^{|\mathcal{F}_i^{(g)}|}$, so the complement gives the contamination probability. In $\mathbf{x}_g^{(\prec i)}$, all future positions are masked by definition, confirming the context mismatch.\hfill$\square$

\smallskip
Thus, random remasking methods evaluate the likelihood ratio under a conditioning distribution that differs from the trajectory-consistent one.

\smallskip
\noindent\textit{Remark.}
Let $r^*_{g,i}(\theta)=\pi_\theta(v_{g,i}\mid\mathbf{x}_g^{(\prec i)},c)/\pi_{\theta_{\mathrm{old}}}(v_{g,i}\mid\mathbf{x}_g^{(\prec i)},c)$ denote the trajectory-consistent ratio, where $\mathbf{x}_g^{(\prec i)}$ is the true denoising state just before token $i$ is revealed (Section~\ref{app:unified_framework}). The claim above concerns the \emph{conditioning distribution}, not a guaranteed pointwise inequality: $\tilde{r}_{g,i}$ and $r^*_{g,i}$ could coincide numerically for specific~$\theta$, but whenever contamination occurs the estimator is structurally inconsistent with the trajectory-consistent objective.

Unlike $\mathbf{x}_g^{(\prec i)}$, $\mathbf{C}_\emptyset$ additionally masks past tokens, so HT-GRPO does not recover $r^*_{g,i}$.
This is a deliberate design choice: exact computation of $r^*_{g,i}$ requires marginalizing over all mask configurations consistent with the generation order, which is combinatorially intractable (Section~\ref{sec:method}).
HT-GRPO instead defines a \emph{unified} estimator using the same context $\mathbf{C}_\emptyset$ for every token:
\begin{equation}
\hat{r}_{g,i}(\theta)
=\frac{\pi_\theta(v_{g,i}\mid\mathbf{C}_\emptyset,c)}{\pi_{\theta_{\mathrm{old}}}(v_{g,i}\mid\mathbf{C}_\emptyset,c)}.
\end{equation}
This estimator has three properties. First, it is contamination-free by construction. Second, it is self-consistent: the conditioning context is identical for both policies, ensuring a well-defined likelihood ratio. Third, it is entropy-preserving: predictive entropy under $\mathbf{C}_\emptyset$ is not smaller in expectation than under contexts that condition on additional structural token information (\hyperref[prop:entropy-lower-bound]{Proposition~C.3}), which maintains reward variance and prevents GRPO advantage collapse.

\phantomsection\label{prop:entropy-lower-bound}
\paragraph{Proposition C.3 (Entropy Lower Bound under Full Masking).}
\textit{Revealing structural tokens reduces uncertainty over refinement tokens.}
Let $\mathbf{S}=\{V_j:j\in\mathcal{S}_{g,\mathrm{structure}}\}$ denote the structural token values and $\mathbf{x}^{\mathrm{structure}}(\mathbf{S})$ the context revealing only those positions. For any refinement token $i\notin\mathcal{S}_{g,\mathrm{structure}}$,
\begin{equation}
H\!\left(\pi_\theta(\cdot \mid \mathbf{C}_\emptyset,c)\right)
\ge
\mathbb{E}_{\mathbf{S}}\!\left[H\!\left(\pi_\theta(\cdot \mid \mathbf{x}^{\mathrm{structure}}(\mathbf{S}),c)\right)\right].
\end{equation}

\noindent\textit{Proof.}
By Assumption~\eqref{eq:consistency}, the left side equals $H(V_i\mid c)$ and the right side equals $H(V_i\mid\mathbf{S},c)$. Non-negativity of $I(V_i;\mathbf{S}\mid c)=H(V_i\mid c)-H(V_i\mid\mathbf{S},c)\ge 0$ gives the result.\hfill$\square$

\smallskip
In practice, once structural layout is fixed, refinement tokens become nearly deterministic: most rollouts look alike, reward variance collapses, and the GRPO advantage shrinks toward zero. This is \textbf{conditional low-entropy degradation} (Section~\ref{sec:method}). Using $\mathbf{C}_\emptyset$ avoids revealing any structural layout, keeping entropy high and preserving the learning signal.

\section{Standard GRPO Objective}
\label{app:grpo_standard}

Given a prompt $q$, GRPO~\cite{deepseekmath} samples $G$ complete outputs $\{o_i\}_{i=1}^{G}$ from the behavior policy $\pi_{\theta_{\mathrm{old}}}$ and maximizes the clipped surrogate objective:
\begin{align}
\label{eq:grpo_standard}
\mathcal{J}_{\mathrm{GRPO}}(\theta)
= \mathbb{E}_{q,\,\{o_i\}\sim\pi_{\theta_{\mathrm{old}}}}\!\Bigg[
  &\frac{1}{G}\sum_{i=1}^{G}\frac{1}{|o_i|}\sum_{t=1}^{|o_i|}
  \min\!\Big(
    r_{i,t}(\theta)\,\hat{A}_{i},\;
    \mathrm{clip}\!\left(r_{i,t}(\theta),1{-}\epsilon,1{+}\epsilon\right)\hat{A}_{i}
  \Big) \notag \\
  &-\;\beta\,\mathbb{D}_{\mathrm{KL}}\!\left(\pi_\theta\,\|\,\pi_{\mathrm{ref}}\right)
\Bigg],
\end{align}
where the importance ratio
\begin{equation}
\label{eq:importance_ratio}
r_{i,t}(\theta)
=
\frac{\pi_\theta\!\left(o_{i,t}\mid q,\,o_{i,<t}\right)}
     {\pi_{\theta_{\mathrm{old}}}\!\left(o_{i,t}\mid q,\,o_{i,<t}\right)}
\end{equation}
measures the relative likelihood of token $o_{i,t}$ under the current policy versus the behavior policy, given the same causal prefix $o_{i,<t}$.
The group-relative advantage
\begin{equation}
\label{eq:advantage}
\hat{A}_i = \frac{R_i - \mathrm{mean}\!\left(\{R_j\}_{j=1}^{G}\right)}{\mathrm{std}\!\left(\{R_j\}_{j=1}^{G}\right) + \delta}
\end{equation}
is a scalar derived from the outcome reward $R_i$ and broadcast uniformly to all tokens in output $o_i$.
In practice, the same sampled batch $\{o_i\}$ is reused for $K$ successive gradient steps, with all importance ratios computed relative to the fixed behavior policy $\pi_{\theta_{\mathrm{old}}}$.

% \begin{figure}[t]
%     \centering
%     \includegraphics[width=1\linewidth]{fig/ablation.pdf}
%     \vspace{-0.22in}
%       \caption{\textbf{(a) Stage organization and budget allocation:} The full Global$\to$Structure$\to$Refinement schedule with a structure-biased 2:4:2 budget consistently outperforms single-stage and two-stage variants. \textbf{(b) Sensitivity to structure ratio $\alpha$:} Performance peaks at $\alpha = 0.3$ and degrades when the boundary is too narrow or too broad. \textbf{(c) Component analysis:} Hierarchical Credit Assignment and Prompt-Conditioned Estimator each provide independent gains, validating both design choices.}
%     \label{fig:ablation}
% \end{figure}

\input{sec/tables/tab_ablate_a3}

\section{Ablation Study Details}
\label{app:ablation}

All ablations run on MMaDA with $K=8$ and $\alpha=0.3$ unless otherwise stated.

\noindent\textbf{A1: Staged RL Training is Necessary for dMLLMs.}
Figure~\ref{fig:ablation}\textcolor{red}{(a)} compares various budget allocation strategies under a fixed inner-loop budget $K=8$.
Both single-stage and two-stage variants yield limited overall performance, scoring only between 78.3 and 80.2.
The full three-stage Sketch-Then-Paint schedule (Global$\to$Structure$\to$Refinement with a 2:4:2 budget) breaks this bottleneck, reaching 83.3 overall, confirming that the coarse-to-fine ordering itself—not merely token-group separation—drives the gain.

\begin{figure}[h]
    \centering
    \includegraphics[width=\linewidth]{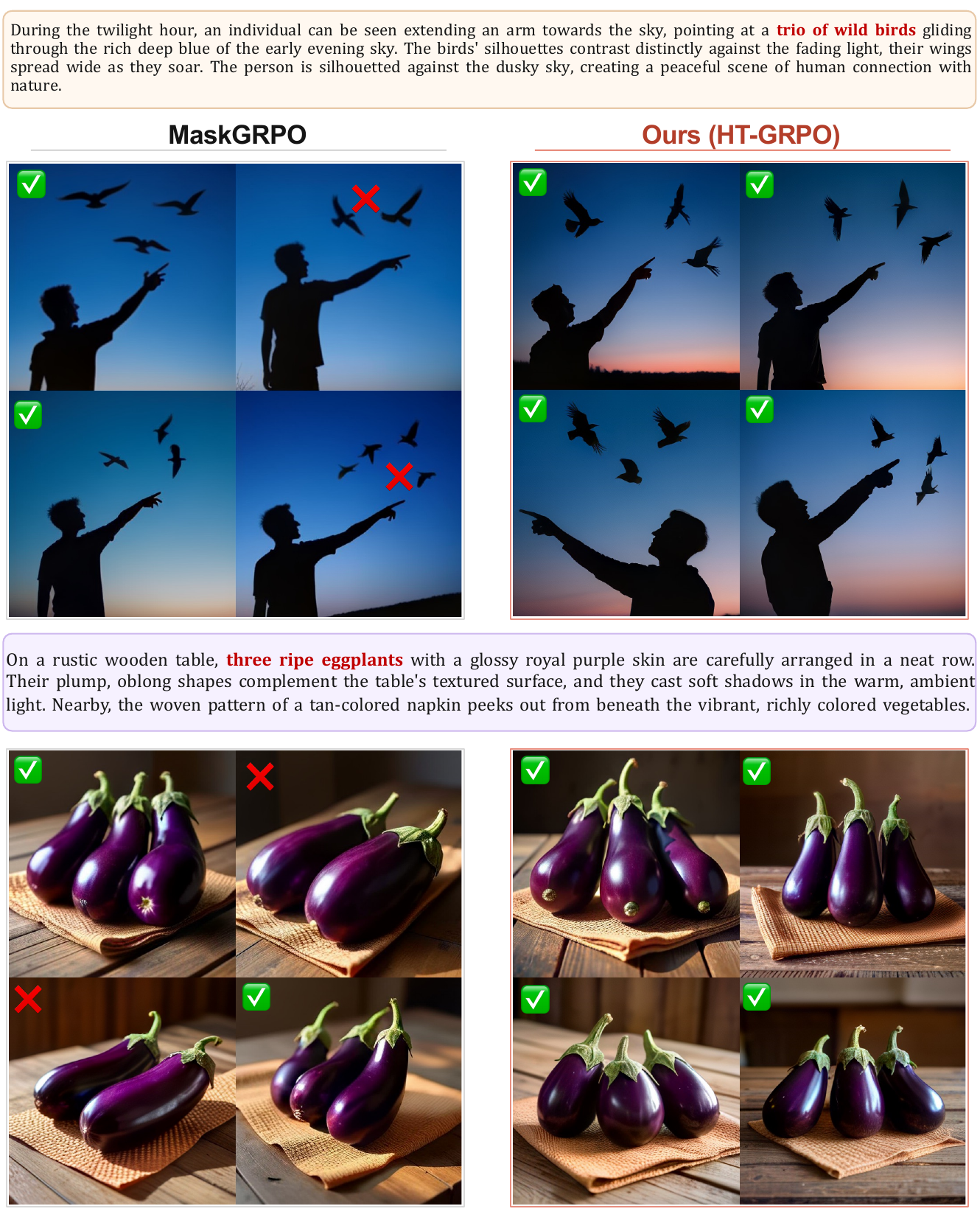}
    \caption{
    \textbf{DPG-Bench counting examples on Lumina-DiMOO.}
    For each prompt, we compare four samples per method: MaskGRPO on the left and HT-GRPO on the right.
    HT-GRPO more consistently preserves the requested object count.
    }
    \label{fig:demo-counting}
\end{figure}

\noindent\textbf{A2: Structure-biased Budget Allocation Maximizes Returns.}
Figure~\ref{fig:ablation}\textcolor{red}{(a)} also explores budget distribution across stages.
The default 2:4:2 allocation achieves the best overall score of 83.3.
Refinement-biased (2:2:4) and global-biased (4:2:2) allocations underperform, indicating that structural tokens require more optimization steps to resolve layout and compositional relations.

\begin{figure}[h]
    \centering
    \includegraphics[width=\linewidth]{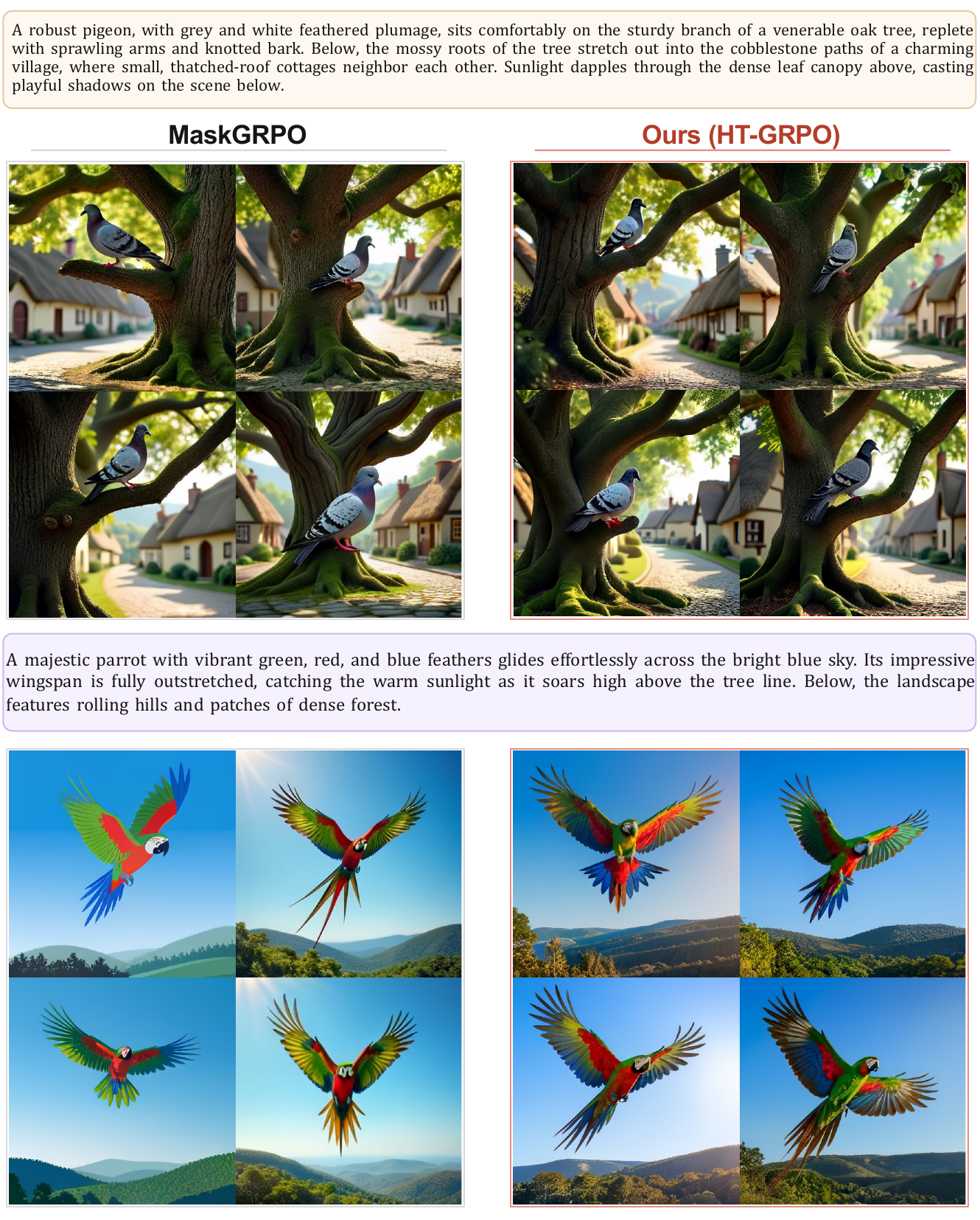}
    \caption{
    \textbf{DPG-Bench scene-level structural completeness on Lumina-DiMOO.}
    Each prompt compares MaskGRPO and HT-GRPO with four samples per method.
    HT-GRPO better preserves relative scale, foreground--background structure, and subject pose across complex scene descriptions.
    }
    \label{fig:demo-structure}
    \vspace{-0.3in}
\end{figure}

\noindent\textbf{A3: Superiority of Coarse-to-fine Annealing Schedule.}
Table~\ref{tab:ablate-a3} evaluates how the annealing schedule controls token sampling within each stage.
Static strategies—constant full coverage ($\gamma_{\max}{=}\gamma_{\min}{=}1.0$, 79.82) and fixed sparse sampling ($\gamma_{\max}{=}\gamma_{\min}{=}0.2$, 80.43)—both underperform dynamic variants.
Ascending annealing ($\gamma{:}\,0.5{\to}1.0$, 82.41) improves over static baselines but lags behind our method: starting with sparse coverage causes unstable gradient directions in early updates.
Our linear decay schedule ($\gamma{:}\,1.0{\to}0.5$) achieves the best overall score of \textbf{83.31}.

\noindent\textbf{A4: Sensitivity to Structure Ratio $\alpha$.}
Figure~\ref{fig:ablation}\textcolor{red}{(b)} evaluates the structural-group boundary.
Performance peaks at $\alpha=0.3$ (83.3). A smaller ratio ($\alpha=0.1$) leaves many layout-relevant tokens under-optimized (80.2), while a larger ratio ($\alpha=0.5$) dilutes the hierarchy by including tokens with rich visual context, both degrading performance.

\noindent\textbf{A5: Ablation on Credit Weighting \& Ratio Conditioning.}
Figure~\ref{fig:ablation}\textcolor{red}{(c)} examines the two components of Section~\ref{subsec:credit} and Section~\ref{subsec:prompt-estimator}.
Discarding credit weights ($\lambda_{\mathrm{s}}=\lambda_{\mathrm{r}}=1$) drops the overall score to 80.8, confirming that amplifying structural-token updates is critical for layout composition.
Replacing $\mathbf{C}_\emptyset$ with revealed structural contexts yields 80.6: once the structural layout is exposed, refinement-token distributions become overly sharp, collapsing ratio variation and weakening the RL signal—precisely the conditional low-entropy degradation identified in \hyperref[prop:entropy-lower-bound]{Proposition~C.3}.

% \subsection{Implementation Supplement}
% \label{app:implementation}

% (old commented block removed)

% \subsubsection{Hyperparameter Settings}
% Table~\ref{tab:appendix-hyperparams} lists all hyperparameters used in our experiments. HT-GRPO is implemented on top of the MaskGRPO~\cite{maskgrpo} codebase, inheriting its optimizer, KL penalty coefficient, and RL loop without modification; only the HT-GRPO-specific settings below are new.

% \input{sec/tables/tab_appendix_hyperparams}

\input{sec/tables/tab_appendix_hyperparams}

\section{Implementation Supplement}
\label{appendix:settings}
\subsection{Random-Subset Scheduling Functions}
\label{app:scheduling}
Let $k_s\in\{0,\ldots,n_s-1\}$ denote the within-stage update counter for stage $s$. The general sampling rate is
\begin{equation}
\gamma_{k_s}^{(s)} = \gamma_{\min}^{(s)} + \bigl(\gamma_{\max}^{(s)}-\gamma_{\min}^{(s)}\bigr)\cdot\phi\!\left(\frac{k_s}{\max(1,n_s-1)}\right),
\end{equation}
where $\phi:[0,1]\to[0,1]$ is a shape function.
The main text uses $\phi(p)=1{-}p$ (\texttt{down} mode), which specializes to Eq.~\eqref{eq:schedule}:
\begin{equation}
\gamma_{k_s}^{(s)} = \gamma_{\min}^{(s)} + \bigl(\gamma_{\max}^{(s)}-\gamma_{\min}^{(s)}\bigr)\cdot\frac{\max(1,n_s-1)-k_s}{\max(1,n_s-1)}.
\end{equation}
We additionally support the following variants (Table~\ref{tab:appendix-schedule}):

\input{sec/tables/tab_appendix_schedule}

Each stage maintains its own $\gamma_{\max}^{(s)}$ and $\gamma_{\min}^{(s)}$, allowing different coverage ranges for the global, structural, and refinement stages.

\subsection{Hyperparameter Settings}
Table~\ref{tab:appendix-hyperparams} lists all hyperparameters used in our experiments. HT-GRPO is implemented on top of the MaskGRPO codebase, inheriting its optimizer, KL penalty coefficient, and RL loop without modification; only the HT-GRPO-specific settings below are new.

\begin{figure}[h]
    \centering
    \includegraphics[width=\linewidth]{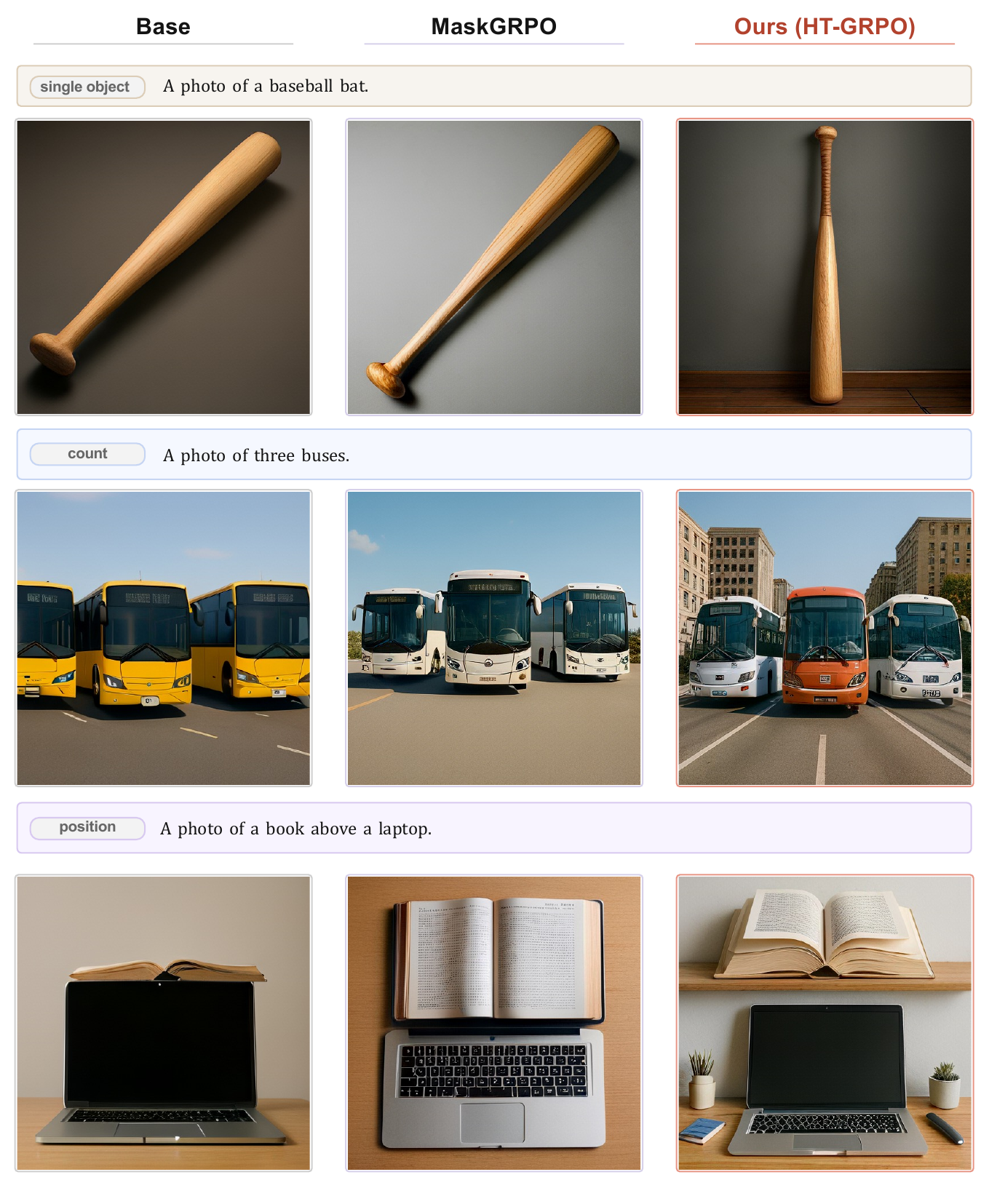}
    \vspace{-0.25in}
    \caption{
    \textbf{GenEval qualitative comparison.}
    From left to right: base model, MaskGRPO, and HT-GRPO.
    HT-GRPO improves visual fidelity, object counting, and spatial relation grounding while producing more natural scene compositions.
    }
    \label{fig:demo-geneval}
\end{figure}

\input{sec/alg_htgrpo}
\section{Qualitative Results}
\label{app:qualitative}
\subsection{DPG-Bench: Counting Accuracy}
Figure~\ref{fig:demo-counting} compares MaskGRPO and HT-GRPO on two DPG-Bench prompts that require reliable counting on Lumina-DiMOO.
The first prompt requires a person pointing toward a trio of birds, where the model must preserve both the object count and the human--object spatial relation.
The second prompt requires exactly three eggplants arranged on a table.
MaskGRPO often produces an incorrect number of target objects, while HT-GRPO more consistently satisfies the requested count.

This illustrates two benefits of the Sketch-Then-Paint hierarchy.
First, counting is treated as a structural decision rather than a refinement detail.
HT-GRPO updates all tokens in the Global stage and then focuses on structural tokens in the Structure stage, allowing the model to determine how many objects should appear and where they should be placed before local appearance is refined.
Second, the staged inner loop prevents counting-related structural decisions from being diluted by refinement updates.
MaskGRPO can mix structure and texture tokens in the same update, so the learning signal for object count and spatial placement competes with color and surface-detail optimization.
HT-GRPO separates these roles by optimizing structure before refinement, making the correct count and layout more stable across samples.

\subsection{DPG-Bench: Scene-Level Structural Completeness}

Figure~\ref{fig:demo-structure} compares MaskGRPO and HT-GRPO on two DPG-Bench prompts that require coherent scene-level structure and foreground--background composition.
For the first prompt, both methods can generate the pigeon, oak tree, and village background, but MaskGRPO shows weaker spatial grounding: the tree roots are not always naturally connected to the ground, and the foreground tree can appear flattened against the village backdrop.
The pigeon is also sometimes disproportionately large relative to the tree and cottages, further weakening the scene hierarchy.
HT-GRPO better preserves the spatial relationship among the pigeon, oak branches, tree trunk, roots, and village background, producing a more coherent foreground--background structure.

For the second prompt, which requires a parrot with fully outstretched wings flying above hills and forest, MaskGRPO can generate the target bird but often renders it with a flatter, less three-dimensional foreground structure.
Its wing geometry and flying pose are less naturally integrated with the landscape.
HT-GRPO more reliably preserves the outstretched wing structure, flying pose, and background scene context, resulting in a stronger sense of depth and scene-level coherence.
These examples suggest that the Sketch-Then-Paint hierarchy helps establish global scene layout and object configuration before local appearance refinement.

\subsection{GenEval: Spatial Grounding and Visual Fidelity}

Figure~\ref{fig:demo-geneval} compares the base model, MaskGRPO, and HT-GRPO on representative GenEval examples covering single-object generation, object counting, and spatial relation understanding.
Beyond improving correctness, HT-GRPO also produces more natural and visually diverse images.
In the single-object example, all methods generate a baseball bat, while HT-GRPO renders a cleaner object with a more realistic scene composition.
In the counting example, HT-GRPO preserves the requested three buses while producing a more diverse urban scene.
The spatial-relation example is particularly illustrative: MaskGRPO tends to entangle the book and laptop into a single top-down composition, making the ``book above laptop'' relation ambiguous.
In contrast, HT-GRPO separates the two objects into a clear vertical arrangement, placing the book above the laptop and enriching the scene with natural contextual elements.
This suggests that HT-GRPO improves not only object-level accuracy, but also spatial grounding and aesthetic scene construction.

\section{Complete Training Algorithm}
We propose HT-GRPO Training in Algorithm~\ref{alg:htgrpo}.
For completeness, the entire training pipeline is summarized in a step-by-step form.

%% file: sec/5_related_work.tex
\section{Related Work}
\subsection{Diffusion Multi-Modal Large Language Models}
Discrete diffusion modeling is rapidly emerging as a highly promising paradigm, showing great potential to replace the traditional autoregressive (AR) modeling.
Early discrete diffusion methods~\cite{austin2021d3pm,lou2023sedd} built the foundation for this shift by establishing the core principles of token-level denoising.
Building on these methods, diffusion large language models (dLLMs)~\cite{gong2024scalingdlm,nie2025llada,llada1.5} show that text generation can be treated as a parallel and iterative denoising process.
Expanding this framework to multi-modal domains, LLaDA-V~\cite{lladav} used visual instruction tuning to show that diffusion models can effectively combine visual and textual data.
Driven by this progress, recent studies focus on developing unified diffusion multi-modal large language models (dMLLMs)~\cite{mmada,shi2025muddit,luminadimoo,LLaDA2Uni}. These models further extend the capabilities of image generation.
This success also inspires new research on downstream applications for dMLLMs. Key areas include reinforcement learning algorithm design~\cite{maskgrpo,mmada}, image editing~\cite{tian2025mmadaparallel}, test-time scaling~\cite{xin2025dmllm}, and sampling acceleration~\cite{migm-shortcut}.

\subsection{Reinforcement Learning Alignment for dMLLMs}
Applying Reinforcement Learning (RL) to discrete diffusion models introduces unique challenges, primarily because dynamic masking breaks the standard policy gradient assumption.
To bypass this, existing works generally diverge into two directions.
Random remasking methods~\cite{maskgrpo,d1,unigrpo} approximate intermediate states by randomly masking tokens. 
While maintaining the efficiency of a single forward pass, these approximated contexts often deviate from the actual denoising trajectory.
Conversely, trajectory recording methods~\cite{tracerl,cjgrpo,dtreerpro,agrpo} cache the entire denoising process to build a precise Markov Decision Process (MDP).
However, this rigorousness comes at a heavy price, as the computational and memory costs scale linearly with the number of denoising steps.
Crucially, both paradigms share a fundamental flaw: they uniformly assign the final reward to all tokens, ignoring prior insights~\cite{egrpo} that early, high-entropy steps matter more for generation quality.

%% file: sec/tables/tab_appendix_framework.tex
\begin{table}[H]
\centering
\small
\caption{Comparison of three dMLLM RL paradigms under a unified formulation. Limitation rows correspond to the four issues identified in Section~\ref{subsec:limitations}.}
\label{tab:appendix-framework}
\resizebox{\linewidth}{!}{
\begin{tabular}{p{3.2cm}p{3.4cm}p{3.2cm}p{3.2cm}}
\toprule
Dimension & Random Remasking & Trajectory Recording & HT-GRPO \\
\midrule
\multicolumn{4}{l}{\textit{Design choices}} \\[2pt]
Support set $\mathcal{M}_g^{(k)}$ & Remasked positions & All $N$ tokens & Random subset of active stage \\
Conditioning context $\mathbf{C}_{g,i}^{(k)}$ & Randomly retained tokens after remasking & True rollout state $\mathbf{x}_{g}^{(\prec i)}$ & Fully masked state $\mathbf{C}_\emptyset$ \\
Token weight $w_{g,i}$ & Uniform ($=1$) & Uniform ($=1$) & Generation-order-aware ($\lambda_{\mathrm{s}}{>}1,\,\lambda_{\mathrm{r}}{<}1$) \\
\midrule
\multicolumn{4}{l}{\textit{Operational cost}} \\[2pt]
Trajectory storage & Final image only & All $T$ intermediate states & Generation-order ranks $\rho_{g,i}$ \\
Forward passes per inner loop & $1$ & $T$ & $1$ \\
\midrule
\multicolumn{4}{l}{\textit{Limitation analysis}} \\[2pt]
Stage conflation & Yes & No & Resolved \\
Future-token contamination & Yes & No & Eliminated \\
Multi-path coverage & Partial (causally inconsistent) & Limited (single trajectory) & Alleviated (intra-stage sampling) \\
Uniform token reward & Unresolved & Unresolved & Resolved \\
\bottomrule
\end{tabular}
}
\end{table}

%% file: sec/tables/tab_ablate_a3.tex
\begin{table*}[t]
\centering
\small
\caption{\textbf{Ablation study on random-subset annealing}. All experiments use $\alpha=0.3$ and $n_g:n_s:n_r=2:4:2$.}
\label{tab:ablate-a3}
\resizebox{\linewidth}{!}{
\begin{tabular}{llccccccc}
\toprule
Coverage strategy & $(\gamma_{\max},\,\gamma_{\min},\,\phi)$ & Single & Two Obj. & Count & Color & Pos. & Attr. & Overall \\
\midrule
Full coverage         & $(1.0,\;1.0,\;\texttt{constant})$ & 100.00 & 87.88 & 72.50 & \textbf{92.55} & 64.00 & 62.00 & 79.82 \\
Fixed sparse sampling & $(0.2,\;0.2,\;\texttt{constant})$ & 97.50 & 85.86 & 75.00 & 87.23 & 70.00 & \textbf{67.00} & 80.43 \\
Ascending annealing   & $(1.0,\;0.5,\;\texttt{up})$       & 100.00 & 91.92 & 76.25 & 88.30 & 72.00 & 66.00 & 82.41 \\
\midrule
Linear decay (ours)   & $(1.0,\;0.5,\;\texttt{down})$     & \textbf{100.00} & \textbf{92.93} & \textbf{77.50} & 90.43 & \textbf{73.00} & 66.00 & \textbf{83.31} \\
\bottomrule
\end{tabular}
}
\end{table*}

%% file: sec/tables/tab_appendix_hyperparams.tex
\begin{table}[h]
\centering
\small
\caption{\textbf{Hyperparameter settings used in all experiments.}}
\label{tab:appendix-hyperparams}
\begin{tabular}{llc}
\toprule
Category & Hyperparameter & Value \\
\midrule
\multirow{4}{*}{HT-GRPO}
  & Structure ratio $\alpha$                              & $0.3$ \\
  & Stage budget $n_{\mathrm{global}}:n_{\mathrm{structure}}:n_{\mathrm{refinement}}$ & $2:4:2$ \\
  & Structural weight $\lambda_{\mathrm{s}}$          & $1.5$ \\
  & Refinement weight $\lambda_{\mathrm{r}}$          & $0.5$ \\
\midrule
\multirow{3}{*}{Annealing}
  & Schedule mode $\phi$                                  & \texttt{down} \\
  & $\gamma_{\max}$ (all stages)                          & $1.0$ \\
  & $\gamma_{\min}$ (all stages)                          & $0.5$ \\
\midrule
\multirow{3}{*}{RL training}
  & Rollouts per prompt $G$                               & $9$ \\
  & Classifier-free guidance scale                        & $3.5$ \\
  & Reward                                                & HPSv3 + CLIP (ViT-L/14) + UniRwd \\
\midrule
\multirow{1}{*}{Hardware}
  & GPUs                                                  & 8$\times$ A100-80G \\
\bottomrule
\end{tabular}
\end{table}

%% file: sec/tables/tab_appendix_schedule.tex
\begin{table}[h]
\centering
\small
\caption{\textbf{Scheduling functions for random-subset annealing.}}
\label{tab:appendix-schedule}
\resizebox{0.85\linewidth}{!}{
\begin{tabular}{lll}
\toprule
Mode & $\phi(p)$ & Behavior \\
\midrule
\texttt{down} & $1-p$ & Linear decay from high coverage to low coverage \\
\texttt{up} & $p$ & Linear growth from low coverage to high coverage \\
\texttt{constant} & $\frac{1}{2}$ & Fixed coverage independent of progress \\
\bottomrule
\end{tabular}
}
\end{table}

%% file: sec/alg_htgrpo.tex
\begin{algorithm}[h]
    \caption{HT-GRPO Training}
    \label{alg:htgrpo}
    \begin{algorithmic}[1]
    \Require Policy $\pi_\theta$, reference policy $\pi_\mathrm{ref}$, reward function $R(\cdot)$, prompts $\{c\}$
    \Require Structure fraction $\alpha$, stage budgets $n_\mathrm{global}, n_\mathrm{structure}, n_\mathrm{refinement}$
    \Require Annealing rates $\gamma_\mathrm{max}^{(s)}, \gamma_\mathrm{min}^{(s)}$, credit weights $\lambda_{\mathrm{s}}, \lambda_{\mathrm{r}}$, clip ratio $\epsilon$, KL coefficient $\beta$, stability constant $\delta$
    
    \For{each training iteration}
        \State $\theta_\mathrm{old} \leftarrow \theta$
        \Statex \hspace{1.5em}\textit{// Phase 1: Rollout}
        \For{$g = 1, \ldots, G$}
            \State Sample $\mathbf{x}_g^{(0)}$ from $\pi_{\theta_\mathrm{old}}$; record unmasking rank $\rho_{g,i}$ for each token $i$
            \State Compute reward $R_g \leftarrow R(\mathbf{x}_g^{(0)}, c)$
        \EndFor
        \Statex \hspace{1.5em}\textit{// Phase 2: Token Partitioning}
        \State $N_s \leftarrow \lfloor \alpha N \rfloor$
        \State $\mathcal{S}_{g,\mathrm{structure}} \leftarrow \{i \mid \rho_{g,i} \le N_s\}$, \quad $\mathcal{S}_{g,\mathrm{refinement}} \leftarrow \{i \mid \rho_{g,i} > N_s\}$, \quad $\mathcal{S}_{g,\mathrm{global}} \leftarrow \{1,\ldots,N\}$
        \Statex \hspace{1.5em}\textit{// Phase 3: Reward \& Hierarchical Credit Assignment}
        \State $A_g \leftarrow \bigl(R_g - \mathrm{mean}(\{R_j\})\bigr) \;/\; \bigl(\mathrm{std}(\{R_j\}) + \delta\bigr)$
        \State $\tilde{A}_{g,i} \leftarrow \begin{cases} A_g \cdot \lambda_{\mathrm{s}} & i \in \mathcal{S}_{g,\mathrm{structure}} \\ A_g \cdot \lambda_{\mathrm{r}} & i \in \mathcal{S}_{g,\mathrm{refinement}} \end{cases}$
        \Statex \hspace{1.5em}\textit{// Phase 4: Sketch-Then-Paint Staged Optimization}
        \For{$s \in [\mathrm{Global},\, \mathrm{Structure},\, \mathrm{Refinement}]$}
            \For{$k_s = 0, \ldots, n_s - 1$}
                \State $\gamma_{k_s}^{(s)} \leftarrow \gamma_\mathrm{min}^{(s)} + \bigl(\gamma_\mathrm{max}^{(s)} - \gamma_\mathrm{min}^{(s)}\bigr) \cdot \dfrac{\max(1,\, n_s - 1) - k_s}{\max(1,\, n_s - 1)}$
                \For{$g = 1, \ldots, G$}
                    \State Sample $\mathcal{M}_g^{(k_s)} \subseteq \mathcal{S}_{g,s}$ with rate $\gamma_{k_s}^{(s)}$
                \EndFor
                \State Compute $\pi_\theta(v_{g,i} \mid \mathbf{C}_\emptyset, c)$ for all $i \in \bigcup_g \mathcal{M}_g^{(k_s)}$ \Comment{single forward pass}
                \State $r_{g,i}(\theta) \leftarrow \pi_\theta(v_{g,i} \mid \mathbf{C}_\emptyset, c) \;/\; \pi_{\theta_\mathrm{old}}(v_{g,i} \mid \mathbf{C}_\emptyset, c)$
                \State $\mathcal{J}^{(k_s)} \leftarrow \dfrac{1}{G}\!\sum_{g=1}^{G} \dfrac{1}{|\mathcal{M}_g^{(k_s)}|}\!\sum_{i \in \mathcal{M}_g^{(k_s)}} \min\!\bigl(r_{g,i}\,\tilde{A}_{g,i},\;\mathrm{clip}(r_{g,i},1{-}\epsilon,1{+}\epsilon)\,\tilde{A}_{g,i}\bigr) - \beta\,\mathbb{D}_\mathrm{KL}(\pi_\theta \| \pi_\mathrm{ref})$
                \State $\theta \leftarrow \theta + \eta\,\nabla_\theta \mathcal{J}^{(k_s)}$
            \EndFor
        \EndFor
    \EndFor
    \end{algorithmic}
    \end{algorithm}